\documentclass[preprint,12pt,authoryear]{elsarticle}

\usepackage{amssymb}
\usepackage{amsmath}
\usepackage{float}
\usepackage[utf8]{inputenc}
\usepackage[T1]{fontenc}
\usepackage{lmodern}
\usepackage{graphicx}
\usepackage[style=base,figurename=Fig.,labelfont=bf,labelsep=period]{caption}
\usepackage{subcaption}
\usepackage{amsmath}
\usepackage{multirow}
\usepackage{booktabs}
\usepackage{threeparttable}
\usepackage{newtxtext,newtxmath}
\usepackage[colorlinks=true,citecolor=red,linkcolor=black]{hyperref}
\usepackage{placeins}

\journal{Accident Analysis & Prevention}

\begin{document}

\begin{frontmatter}



\title{MDAS-GNN: Multi-Dimensional Spatiotemporal GNN with Spatial Diffusion for Urban Traffic Risk Forecasting} 


\author{Ziyuan Gao}

\affiliation{organization={University College London},
            addressline={Gower Street}, 
            city={London},
            postcode={WC1E 6BT}, 
            state={},
            country={United Kingdom}}

\begin{abstract}
Traffic accidents represent a critical public health challenge, claiming over 1.35 million lives annually worldwide. Traditional accident prediction models treat road segments independently, failing to capture complex spatial relationships and temporal dependencies in urban transportation networks. This study develops MDAS-GNN, a Multi-Dimensional Attention-based Spatial-diffusion Graph Neural Network integrating three core risk dimensions: traffic safety, infrastructure, and environmental risk. The framework employs feature-specific spatial diffusion mechanisms and multi-head temporal attention to capture dependencies across different time horizons. Evaluated on UK Department for Transport accident data across Central London, South Manchester, and SE Birmingham, MDAS-GNN achieves superior performance compared to established baseline methods. The model maintains consistently low prediction errors across short, medium, and long-term periods, with particular strength in long-term forecasting. Ablation studies confirm that integrated multi-dimensional features outperform single-feature approaches, reducing prediction errors by up to 40\%. This framework provides civil engineers and urban planners with advanced predictive capabilities for transportation infrastructure design, enabling data-driven decisions for road network optimization, infrastructure resource improvements, and strategic safety interventions in urban development projects. The source code is released at https://github.com/zygao930/MDAS-GNN
\end{abstract}



\begin{keyword}
traffic accident prediction \sep spatiotemporal graph neural networks \sep multi-dimensional risk assessment \sep urban transportation safety \sep spatial diffusion modeling


\end{keyword}

\end{frontmatter}

\section{Introduction}

Traffic accidents represent one of the most significant public health challenges, claiming approximately 1.35 million lives annually worldwide (\cite{WHO:2018}). In the UK alone, road traffic accidents cause over 1,700 fatalities annually, with associated costs estimated at £16 billion (\cite{DfT:2019}). The complexity of urban transportation systems, characterized by intricate spatiotemporal dependencies and multifaceted risk factors, necessitates sophisticated predictive modeling approaches.

Traditional traffic accident prediction methods have relied on statistical models that treat road segments as independent entities (\cite{Lord:2010a,Mannering:2014a}). However, these approaches fail to capture spatial correlations and temporal dependencies in urban road networks. Recent advances in graph neural networks and spatiotemporal modeling have enabled more accurate traffic safety predictions (\cite{Wu:2020a,Yu:2018a}). However, traffic safety predictions is complicated by heterogeneous contributing factors including traffic patterns, infrastructure characteristics, environmental conditions, and temporal variations (\cite{Cheng:2019a,Zhao:2020a}). Existing models struggle to integrate these diverse risk dimensions while maintaining efficiency and interpretability.

This study proposes MDAS-GNN (Multi-Dimensional Attention-based Spatiotemporal Graph Neural Network), integrating three core risk dimensions: traffic safety risk, infrastructure risk, and environmental risk, through feature-specific spatial diffusion mechanisms and multi-head temporal attention. The approach is validated using UK Department for Transport data (2005-2014) across three metropolitan regions.

The main contributions are: (1) A multi-dimensional feature framework categorizing traffic accident risk factors into three dimensions with tailored spatial diffusion mechanisms; (2) A spatiotemporal graph neural network combining feature-specific spatial attention with weekly-focused temporal modeling; and (3) Comprehensive experimental validation demonstrating superior performance across multiple temporal horizons and geographic regions.

\section{Related Work}
\subsection{Multi-Dimensional Risk Assessment in Traffic Safety}

Traffic accident causation involves complex interactions among human, vehicle, roadway, and environmental factors. Early research adopted single-factor perspectives: traffic volume and speed as exposure metrics (\cite{Lord:2010a}), geometric design and junction characteristics as infrastructure determinants (\cite{Cafiso:2010a}), and weather conditions as environmental modulators (\cite{Edwards:1999a,Andrey:1993a}). While these studies advanced understanding of specific mechanisms, they treat risk factors in isolation rather than as interacting components. Recent research has begun to recognize these complex interactions. For example, an analysis of motorway accident severity in England demonstrates how infrastructure and behavioral factors combine: accidents on the hard shoulder had a fatality rate of 8.38\% compared with 1.77\% on the main carriageway for 2005–2011 (\cite{Michalaki:2015}), with the elevated severity resulting from both infrastructure location and behavioral factors.

Building on such insights, traffic accident risk assessment has evolved toward multi-dimensional frameworks through work by Hadayeghi et al. (\cite{Hadayeghi:2010a}), ensemble methods (\cite{Zhang:2020a}), and hierarchical models (\cite{Cheng:2019a}). However, these approaches often treat different risk factors uniformly without considering their distinct temporal and spatial propagation characteristics. This presents a fundamental challenge: existing studies have identified numerous contributing variables (over fifty in the UK's STATS19 reports (\cite{DfT:2020})) but lack systematic principles for organizing these heterogeneous factors (\cite{Mannering:2014a}). Current approaches face a dilemma between aggregating all variables into a single index (losing interpretability) and modeling each factor independently (creating computational complexity and overfitting risks with sparse data (\cite{Lord:2010a})). While established safety paradigms suggest that risk factors operate on distinct temporal and spatial scales, no predictive framework explicitly leverages this insight.

\subsection{Spatiotemporal Modeling for Accident Prediction}

\subsubsection{Spatial Dependencies and Propagation Mechanisms}

Building on decades of traffic safety research, early traffic accident prediction relied on statistical models like Poisson and negative binomial regression (\cite{Lord:2010a,Mannering:2014a}), which assume spatial independence and struggle with temporal dependencies. Machine learning approaches including SVMs (\cite{Kumar:2013}) and random forests (\cite{Chang:2005}) showed improvements but remain limited in capturing spatiotemporal relationships. Recent deep learning approaches applied CNNs (\cite{Ren:2018}) and RNNs (\cite{Wang:2019}) to traffic safety prediction, but treat spatial relationships as Euclidean, failing to capture road network topology.

Traditional spatial analysis methods in traffic safety include Geographically 
Weighted Regression (GWR), which captures spatial heterogeneity through 
location-varying coefficients but assumes Euclidean distance relationships 
(\cite{Xu:2015a}), and spatial clustering techniques that identify accident 
hot-spots through kernel density estimation or spatial scan statistics 
(\cite{Zhao:2020a,Cheng:2019a}). While GWR models reveal regional variations 
in risk factors, they ignore road network topology and directional traffic 
flows. Hot-spot identification methods excel at visualization and resource 
allocation but provide limited predictive capability for understanding how 
risk propagates from one location to another.

To address these spatial modeling limitations, graph neural networks have revolutionized spatiotemporal modeling in transportation. These approaches leverage spatial diffusion principles grounded in Tobler's First Law (\cite{Tobler:1970a}), with transportation networks showing strong spatial dependencies (\cite{Wang:2006a}). Li et al. (\cite{Li:2018a}) introduced DCRNN for traffic flow prediction, combining graph convolution with recurrent architectures. Yu et al. (\cite{Yu:2018a}) proposed STGCN using temporal convolutions for better efficiency. Attention mechanisms further enhanced these networks: ASTGCN (\cite{Guo:2019a}) dynamically weights spatial and temporal dependencies, while GMAN (\cite{Zheng:2020}) applies multi-level attention. However, these models primarily focus on traffic flow prediction with limited adaptation to traffic safety applications.

\subsubsection{Temporal Dynamics and Multi-Scale Patterns}

Accident risk exhibits temporal variations across multiple scales, from daily 
weather fluctuations to decadal infrastructure evolution. Traditional time-series 
methods (ARIMA, exponential smoothing) assume stationarity and struggle with 
irregular safety patterns (\cite{Lord:2010a}). Machine learning approaches including 
SVMs (\cite{Kumar:2013}) and random forests (\cite{Chang:2005}) showed improvements 
but remain limited in capturing temporal dependencies. Recent deep learning 
methods applied CNNs (\cite{Ren:2018}) and RNNs (\cite{Wang:2019}) to traffic safety 
prediction, demonstrating improved capability in modeling complex temporal patterns.

Building on these advances, spatiotemporal GNNs integrate spatial graph 
structure with temporal modeling through recurrent units (e.g., DCRNN 
(\cite{Li:2018a})), temporal convolutions (e.g., STGCN (\cite{Yu:2018a})), 
or attention mechanisms (e.g., ASTGCN (\cite{Guo:2019a}), GMAN 
(\cite{Zheng:2020})). These models have achieved strong performance for 
traffic flow prediction, where weekly aggregation has emerged as optimal 
for balancing signal-to-noise ratio against temporal resolution 
(\cite{Smith:2019,Batty:2013}). However, their application to traffic 
safety scenarios remains limited.

The critical limitation is that existing temporal models apply uniform temporal 
receptive fields to all risk factors, ignoring fundamental differences in how 
different risk dimensions evolve over time. Previous studies have documented 
that environmental conditions (weather, surface states) exhibit high short-term 
volatility requiring immediate responses (\cite{Edwards:1999,Andrey:1993}), 
infrastructure characteristics change at moderate rates with tactical improvement 
cycles (\cite{Cafiso:2010}), while structural traffic patterns display long-term 
stability with strategic redesign horizons (\cite{Lord:2010a}). However, existing 
spatiotemporal models treat these factors uniformly with fixed attention windows, 
forcing models to learn compromise representations that optimally serve none of 
the dimensions.

\subsection{Spatial Diffusion in Accident Analysis}

Traffic accidents exhibit systematic spatial clustering due to shared risk factors and connected road network effects, violating the independence assumptions of traditional prediction models. This spatial autocorrelation requires explicit modeling of how safety conditions propagate between neighboring road segments to achieve accurate risk assessment (\cite{Black:1998, Flahaut:2003}). The theoretical foundation rests on Tobler's First Law of Geography, which establishes that spatial proximity influences correlation strength in geographic phenomena, with transportation networks demonstrating particularly strong spatial dependencies due to traffic flow continuity and shared infrastructure characteristics \cite{Wang:2006}.

Spatial diffusion approaches have evolved from basic autocorrelation techniques to sophisticated network-based models. Early applications employed Geographically Weighted Regression to capture spatial heterogeneity in crash patterns (\cite{Hadayeghi:2010}), while spatial econometric methods introduced explicit spatial lag and error terms for crash frequency modeling. Aguero-Valverde and Jovanis (\cite{Aguero:2006}) demonstrated significant improvements using spatial lag models for crash frequency prediction, establishing the importance of neighborhood effects in safety analysis. Recent graph neural network applications have incorporated network topology into spatial modeling frameworks. Zhou et al. (\cite{Zhou:2020}) and Chen et al. (\cite{Chen:2021} )developed graph convolutional approaches for spatiotemporal accident prediction, demonstrating improved performance over traditional methods by respecting road network connectivity rather than Euclidean distance relationships.

However, current spatial diffusion methods apply uniform propagation mechanisms across all risk factors. Research in spatial epidemiology and environmental criminology suggests that different phenomena exhibit distinct spatial correlation patterns based on their underlying transmission mechanisms (\cite{Cliff:1981}). Infrastructure-related risks typically demonstrate localized clustering due to the discrete nature of physical facilities, while environmental factors show broader regional correlation patterns reflecting meteorological and geographic processes. Traffic safety risks exhibit intermediate spatial dependencies driven by behavioral and flow-based propagation along connected corridors. Existing approaches treat these different risk dimensions uniformly, which may limit the interpretability of spatial relationships in transportation safety modeling.

\section{Three-Dimensional Risk Framework}

\subsection{Framework Design}

To address the challenge of systematically organizing heterogeneous risk factors, 
we propose organizing accident causation factors into three dimensions distinguished 
by their temporal volatility patterns and spatial propagation characteristics. 
This framework draws on three well-established safety paradigms that collectively 
emphasize the multi-scale nature of accident risk.

The Haddon Matrix (\cite{Haddon1980}) highlights structural exposures (geometry, 
traffic volume) that exhibit stable, long-term patterns and call for strategic 
interventions. The Safe System Approach (\cite{Wegman:2008}) emphasizes modifiable 
protection features (crossings, lighting, junction control) that show medium-term 
variation and can be improved through tactical upgrades. Vision Zero 
(\cite{Johansson:2009}) focuses on dynamic environmental factors (weather, 
surface states) that display high short-term volatility and require immediate 
operational responses. These paradigms converge on a key insight: accident risk 
emerges from factors with distinct temporal volatility patterns and spatial 
propagation behaviors.

Accordingly, we operationalize this into three complementary dimensions : traffic safety risk (stable long-term patterns with 
strong spatial clustering), infrastructure risk (medium-term patterns with 
moderate spatial correlation), and environmental risk (high-volatility short-term 
patterns with regional propagation). This dimensionalization enables (1) 
feature-specific spatial diffusion mechanisms matching distinct propagation scales; 
(2) multi-head temporal attention capturing different volatility patterns; 
(3) actionable decomposition for targeted interventions aligned with planning horizons.

\subsection{Empirical Validation}

We validated this framework using UK Department for Transport accident data 
(2009 to 2014, $N$=899,987) (\cite{DfT:2020}). Four empirical analyses confirm 
that the three dimensions exhibit distinct measurable behaviors (Table~\ref{tab:validation}). 
First, cross-dimensional correlations are low (mean $|r|$=0.095 for traffic, 
0.134 for infrastructure, 0.107 for environmental), confirming orthogonality 
and nonredundant variance capture. Second, environmental risk shows 2.7$\times$ 
higher weekly volatility (coefficient of variation=69.0\%) than traffic safety 
risk (25.6\%), and week-to-week autocorrelation ranges from 0.551 (traffic) to 
0.971 (infrastructure), reflecting distinct temporal stability patterns. Third, 
a 1 km grid analysis of Central London (2,426 cells) shows that traffic safety 
risk is the most spatially clustered (ICC=0.327, corridor-like patterns), 
infrastructure risk is moderately localized (ICC=0.154, junction concentration), 
and environmental risk is the most spatially uniform (ICC=0.091), consistent 
with regional meteorological influence (\cite{Tobler:1970a,Wang:2006a}). Fourth, 
hierarchical regression shows that each dimension adds incremental predictive 
value: combined models raise explained variance from $R^2$=0.080 to 0.096 (+19\%), 
with environmental risk contributing the largest marginal gain (+15.0\%). 
Together, these findings confirm that the three dimensions differ systematically 
in correlation structure, temporal volatility, spatial clustering, and predictive 
contribution, validating the multidimensional premise.

\begin{center}
\captionof{table}{Empirical Validation of Three-Dimensional Risk Framework}
\label{tab:validation}
\footnotesize
\begin{tabular}{l l c c c}
\hline\hline
\multicolumn{1}{c}{Analysis} & 
\multicolumn{1}{c}{Metric} & 
\multicolumn{1}{c}{Traffic Safety} & 
\multicolumn{1}{c}{Infrastructure} & 
\multicolumn{1}{c}{Environmental} \\
\multicolumn{1}{c}{(1)} & 
\multicolumn{1}{c}{(2)} & 
\multicolumn{1}{c}{(3)} & 
\multicolumn{1}{c}{(4)} & 
\multicolumn{1}{c}{(5)} \\
\hline
\multirow{2}{*}{\parbox{2cm}{\begin{tabular}[t]{@{}l@{}}Dimensional\\Independence\end{tabular}}} 
& Cross correlation & 0.095 & 0.134 & 0.107 \\
& (mean $|r|$) & & & \\
\cline{2-5}
\multirow{3}{*}{\parbox{2cm}{\begin{tabular}[t]{@{}l@{}}Temporal\\Differentiation\end{tabular}}} 
& Coeff. of variation (\%) & 25.6 & 49.5 & 69.0 \\
& Week-to-week autocorr. & 0.551 & 0.971 & 0.664 \\
& Intervention timescale & 2 to 5 yrs & 6 to 18 mo & Immediate \\
\cline{2-5}
\multirow{2}{*}{\parbox{3cm}{\begin{tabular}[t]{@{}l@{}}Spatial\\Propagation\end{tabular}}} 
& Intraclass corr. (ICC) & 0.327 & 0.154 & 0.091 \\
& Between grid var. (\%) & 32.7 & 15.4 & 9.1 \\
\cline{2-5}
\multirow{2}{*}{\parbox{3cm}{\begin{tabular}[t]{@{}l@{}}Predictive\\Complementarity\end{tabular}}} 
& Incremental $R^2$ & 0.080 & +0.003 & +0.012 \\
& Relative improv. (\%) & Base & +3.8 & +15.0 \\
\hline\hline
\end{tabular}
\normalsize

\vspace{0.5em}
\begin{minipage}{\textwidth}
\footnotesize
Note: $N$=899,987 accidents (UK, 2009 to 2014); Spatial analysis: London, 2,426 grid cells (1km)
\end{minipage}
\end{center}

\section{Methodology}

\subsection{Data}
This study utilizes the UK Department of Transport's comprehensive traffic accident dataset spanning 2005--2014, containing over 1.6 million police-reported accidents. The main analysis focuses on Central London (2009--2014), while experiments are conducted across three UK metropolitan regions to validate generalizability. For studies in London, the network extraction method models 2,144 road segments as graph nodes positioned at their geometric centroids, connected by 9,612 spatial links(as shown in Fig.\ref{fig:spatialtemporal}(a) and (b)), forming a road-segment-based topology that captures the continuous nature of incident risk across road corridors with 39,820 total accidents.

\subsection{Temporal Aggregation Strategy}

The choice of temporal aggregation granularity critically impacts both data quality and model performance in spatiotemporal graph neural networks (\cite{Li:2018a,Wu:2020a}). To adopt the best temporal aggregation, the analysis is conducted on the raw accident intensity feature data from 2012-2014 as a representative sample to illustrate the aggregation effects, as shown in Fig.\ref{fig:spatialtemporal}(c). 

Weekly aggregation achieves superior noise reduction, improving the signal-to-noise ratio from 2.888 (daily) to 5.437 (weekly)—an 88.3\% improvement through temporal averaging that preserves underlying traffic safety patterns while attenuating random fluctuations (\cite{Smith:2019a}). Additionally, weekly cycles align with human behavioral patterns and policy schedules, providing more stable temporal dependencies than daily fluctuations (\cite{Batty:2013a}). While monthly aggregation further improves SNR to 7.727, it over-smooths temporal variations and loses critical short-term dynamics necessary for accurate traffic safety modeling. This aggregation strategy is validated across all feature dimensions, as both the raw accident data and the derived risk metrics are similarly influenced by underlying temporal noise patterns.

\begin{center}
    \includegraphics[width=\textwidth]{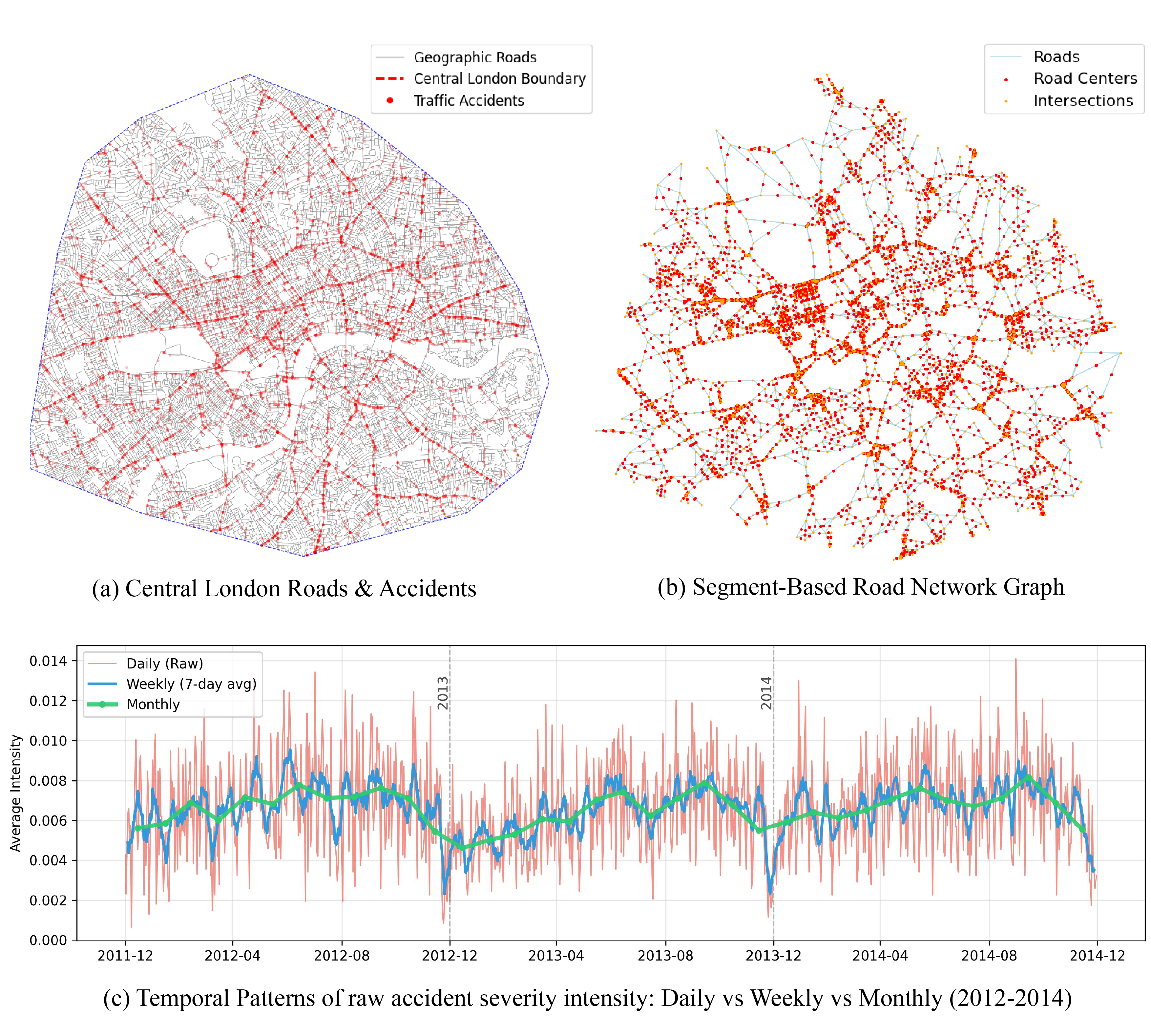}
    \captionof{figure}{Spatiotemporal network representation and temporal aggregation analysis, Sub-figure (a) illustrates road segment graph topology, Sub-figure (b) illustrates Spatial connectivity structure, Sub-figure (c) highlights Signal-to-noise ratio comparison across daily, weekly, and monthly aggregation granularities}
    \label{fig:spatialtemporal}
\end{center}

\subsection{Spatiotemporal Feature Learning Framework}

In complex urban transportation systems, traffic accidents exhibit significant spatiotemporal dependencies and nonlinear dynamic characteristics. Research has consistently demonstrated that traffic accidents are not randomly distributed in space and time, but rather exhibit strong clustering patterns influenced by various environmental, infrastructure, and temporal factors (\cite{Cheng:2019a,Zhao:2020a}). To capture these multi-scale spatiotemporal coupling effects, this study constructs a tensor decomposition-based spatiotemporal feature learning framework.

The temporal sequence is defined as $T = \{t_1, t_2, \ldots, t_W\}$, where $W$ represents the total number of time windows. The spatial node set is denoted as $V = \{v_1, v_2, \ldots, v_N\}$, where $N$ is the total number of road network nodes.

\subsubsection{Multi-dimensional Feature Tensor Construction}

This research defines a multi-dimensional feature tensor $\mathcal{X} \in \mathbb{R}^{W \times N \times F}$, where $F$ represents the feature dimensionality. This tensor-based representation enables the integration of heterogeneous traffic factors into a unified framework, following the multi-view learning paradigm in traffic analysis (\cite{Zhang:2020a}).

Given the complexity of traffic accidents and their multifaceted nature as identified by comprehensive accident analysis studies (\cite{Mannering:2014a}), the framework designs three core feature dimensions:

\begin{equation}
\mathcal{X}_{t,i,f} = \begin{cases} 
\mathcal{S}_{t,i} & \text{if } f = 0 \text{ (Traffic Safety Risk)} \\
\mathcal{I}_{t,i} & \text{if } f = 1 \text{ (Infrastructure Risk)} \\
\mathcal{E}_{t,i} & \text{if } f = 2 \text{ (Environmental Risk)}
\end{cases}
\end{equation}

The proposed three-feature framework aligns traffic safety risk with distinct temporal intervention scales. Traffic safety risk represents long-term structural hazards that necessitate strategic planning and infrastructural redesign. Safety infrastructure risk reflects medium-term, modifiable elements that can be addressed through tactical improvements, such as upgrades to signage or barriers. Environmental risk captures short-term, dynamic condition that require real-time operational responses. By decomposing risk into these categories, the spatiotemporal graph neural network can model the complex interactions among static design features, semi-static safety systems, and transient environmental factors.

\subsubsection{Traffic Safety Risk Modeling}
The traffic safety risk $\mathcal{S}_{t,i}$ is computed through a dynamic spatiotemporal weighting mechanism that incorporates both casualty impact and temporal clustering effects:
\begin{equation}
\mathcal{S}_{t,i} = \sum_{k} \left[ \log(C_k + 1) \cdot \omega_{\text{sev}}(s_k) \cdot \omega_{\text{temp}}(t,t_k) \right]
\end{equation}
where $C_k$ denotes the number of casualties in the $k$-th accident. The logarithmic transformation is commonly used in accident severity modeling to handle the skewed distribution of casualty counts (\cite{Lord:2010a}). 

The comprehensive severity weight $\omega_{\text{sev}}(s_k)$ integrates multiple risk factors through a multiplicative framework:
\begin{equation}
\omega_{\text{sev}}(s_k) = \omega_{\text{severity}}(s_k) \cdot \omega_{\text{road}}(r_i) \cdot \omega_{\text{speed}}(v_i)
\end{equation}
where $\omega_{\text{severity}}(s_k)$ differentiates between accident types based on their societal impact and emergency response requirements, $\omega_{\text{road}}(r_i)$ reflects inherent basic potential risk associated with different road configurations (as shown in Table \ref{table:severity_road_weights}), and $\omega_{\text{speed}}(v_i) = 0.5 + v_i/120$ normalizes speed limits to risk factors, with higher speed limits resulting in increased traffic risks.

\begin{center}
\captionof{table}{Severity level and road context risk}
\label{table:severity_road_weights}
\footnotesize
\begin{tabular}{l l c c}
\hline\hline
\multicolumn{1}{c}{Factor} & 
\multicolumn{1}{c}{Category} & 
\multicolumn{1}{c}{Weight} & 
\multicolumn{1}{c}{Risk Level} \\
\multicolumn{1}{c}{(1)} & 
\multicolumn{1}{c}{(2)} & 
\multicolumn{1}{c}{(3)} & 
\multicolumn{1}{c}{(4)} \\
\hline
\multirow{3}{*}{\parbox{3cm}{\begin{tabular}[t]{@{}l@{}}Severity Level\end{tabular}}} 
& 3 (Slight) & 1.0 & Minor injury accidents \\
& 2 (Serious) & 2.0 & Serious injury accidents \\
& 1 (Fatal) & 3.0 & Most severe accidents \\
\cline{2-4}
\multirow{5}{*}{\parbox{3cm}{\begin{tabular}[t]{@{}l@{}}Road Context Risk\end{tabular}}}
& Single carriageway & 1.0 & Standard risk \\
& One way street & 1.1 & Slightly elevated risk \\
& Dual carriageway & 1.2 & Moderate-high risk \\
& Slip road & 1.3 & High risk \\
& Roundabout & 1.5 & Highest risk \\
\hline\hline
\end{tabular}
\normalsize
\end{center}

\subsubsection{Infrastructure and Environmental Risk Modeling}

The infrastructure risk $\mathcal{I}_{t,i}$ and environmental risk $\mathcal{E}_{t,i}$ are computed through multivariate conditional probability models, building upon comprehensive risk factor analysis frameworks in transportation safety literature (\cite{Hadayeghi:2010a}):

\begin{equation}
\mathcal{I}_{t,i} = \frac{1}{M} \sum_{m=1}^{M} \omega_{\text{infra}}^{(m)}(\mathbf{c}_{t,i}^{(m)})
\end{equation}

\begin{equation}
\mathcal{E}_{t,i} = \frac{1}{L} \sum_{l=1}^{L} \omega_{\text{env}}^{(l)}(\mathbf{e}_{t,i}^{(l)})
\end{equation}

where $\mathbf{c}_{t,i}^{(m)}$ and $\mathbf{e}_{t,i}^{(l)}$ represent the infrastructure condition vectors and environmental condition vectors, respectively. The weighting functions $\omega_{\text{infra}}^{(m)}$ and $\omega_{\text{env}}^{(l)}$ are derived from empirical risk analysis studies.

Infrastructure risk is evaluated using four factors ($M=4$)  (as shown in Table \ref{table:infrastructure_risk}): pedestrian control measures, crossing facilities, lighting, and junction control, following established assessment frameworks (\cite{Cafiso:2010a}). Environmental risk includes two factors ($L=2$) (as shown in Table \ref{table:environmental_risk}): road surface and weather conditions, consistent with traffic safety models (\cite{Edwards:1999a,Andrey:1993a}).

To observe the temporal evolution pattern of each feature, the temporal evolution plot (as shown in Fig.\ref{fig:three_feature}) reveals longitudinal patterns in traffic safety risk dynamics across 2009-2014, where the green area indicates that the difference between the number of accidents occurring on the same week is similar. It is clear that different features exhibit distinct time series trend. Therefore, this analysis demonstrates that each risk component follows distinct temporal patterns, indicating that predictive models should incorporate feature-specific dynamics rather than assuming uniform temporal behavior across all factors.

\begin{table}[htbp]
\centering
\caption{Infrastructure risk weights}
\label{table:infrastructure_risk}
\footnotesize
\begin{tabular}{p{2.5cm} p{5.5cm} c c}
\hline\hline
\multicolumn{1}{c}{\textbf{Factor}} & 
\multicolumn{1}{c}{\textbf{Category}} & 
\multicolumn{1}{c}{\textbf{Weight}} & 
\multicolumn{1}{c}{\textbf{Risk Level}} \\
\multicolumn{1}{c}{(1)} & 
\multicolumn{1}{c}{(2)} & 
\multicolumn{1}{c}{(3)} & 
\multicolumn{1}{c}{(4)} \\
\hline
\multirow{3}{2.5cm}{Pedestrian \\ Crossing \\ Human Control}
& Control by school crossing patrol & 0.2 & Very low risk \\
& Control by other authorised person & 0.3 & Low risk \\
& None within 50 metres & 0.4 & Moderate risk \\
\cline{2-4}
\multirow{6}{2.5cm}{Pedestrian \\ Crossing \\ Physical Facilities}
& Footbridge or subway & 0.1 & Lowest risk \\
& Pedestrian phase at traffic signal junction & 0.2 & Very low risk \\
& Non-junction pedestrian crossing & 0.3 & Low risk \\
& Zebra crossing & 0.35 & Moderate-low risk \\
& Central refuge & 0.4 & Moderate risk \\
& No physical crossing within 50 meters & 0.6 & High risk \\
\cline{2-4}
\multirow{5}{2.5cm}{Light \\ Conditions}
& Daylight: Street light present & 0.2 & Lowest risk \\
& Darkness: Street lights present and lit & 0.4 & Low risk \\
& Darkness: Street lighting unknown & 0.6 & Moderate risk \\
& Darkness: Street lights present but unlit & 0.7 & High risk \\
& Darkness: No street lighting & 0.8 & Highest risk \\
\cline{2-4}
\multirow{4}{2.5cm}{Junction \\ Control}
& Authorised person & 0.2 & Lowest risk \\
& Automatic traffic signal & 0.3 & Low risk \\
& Stop Sign & 0.5 & Moderate risk \\
& Give way or uncontrolled & 0.7 & High risk \\
\hline\hline
\end{tabular}
\normalsize
\end{table}

\newpage
\begin{center}
\captionof{table}{Environmental risk weights}
\label{table:environmental_risk}
\footnotesize
\renewcommand{\arraystretch}{1}
\begin{tabular}{l l c c}
\hline\hline
\multicolumn{1}{c}{Factor} & 
\multicolumn{1}{c}{Category} & 
\multicolumn{1}{c}{Weight} & 
\multicolumn{1}{c}{Risk Level} \\
\multicolumn{1}{c}{(1)} & 
\multicolumn{1}{c}{(2)} & 
\multicolumn{1}{c}{(3)} & 
\multicolumn{1}{c}{(4)} \\
\hline
\multirow{4}{*}{\parbox{3cm}{\begin{tabular}[t]{@{}l@{}}Road Surface \\ Conditions\end{tabular}}}
& Dry & 0.2 & Lowest risk \\
& Wet/Damp & 0.5 & Moderate risk \\
& Snow & 0.7 & High risk \\
& Flood (Over 3cm of water) & 0.7 & High risk \\
& Frost/Ice & 0.8 & Highest risk \\
\cline{2-4}
\multirow{7}{*}{\parbox{3cm}{\begin{tabular}[t]{@{}l@{}}Weather \\ Conditions\end{tabular}}}
& Fine without high winds & 0.2 & Lowest risk \\
& Fine with high winds & 0.3 & Low risk \\
& Raining without high winds & 0.5 & Moderate risk \\
& Fog or mist & 0.6 & Moderate-high risk \\
& Raining with high winds & 0.7 & High risk \\
& Snowing without high winds & 0.7 & High risk \\
& Snowing with high winds & 0.8 & Highest risk \\
\hline\hline
\end{tabular}
\normalsize
\end{center}

\begin{center}
    \includegraphics[width=\textwidth]{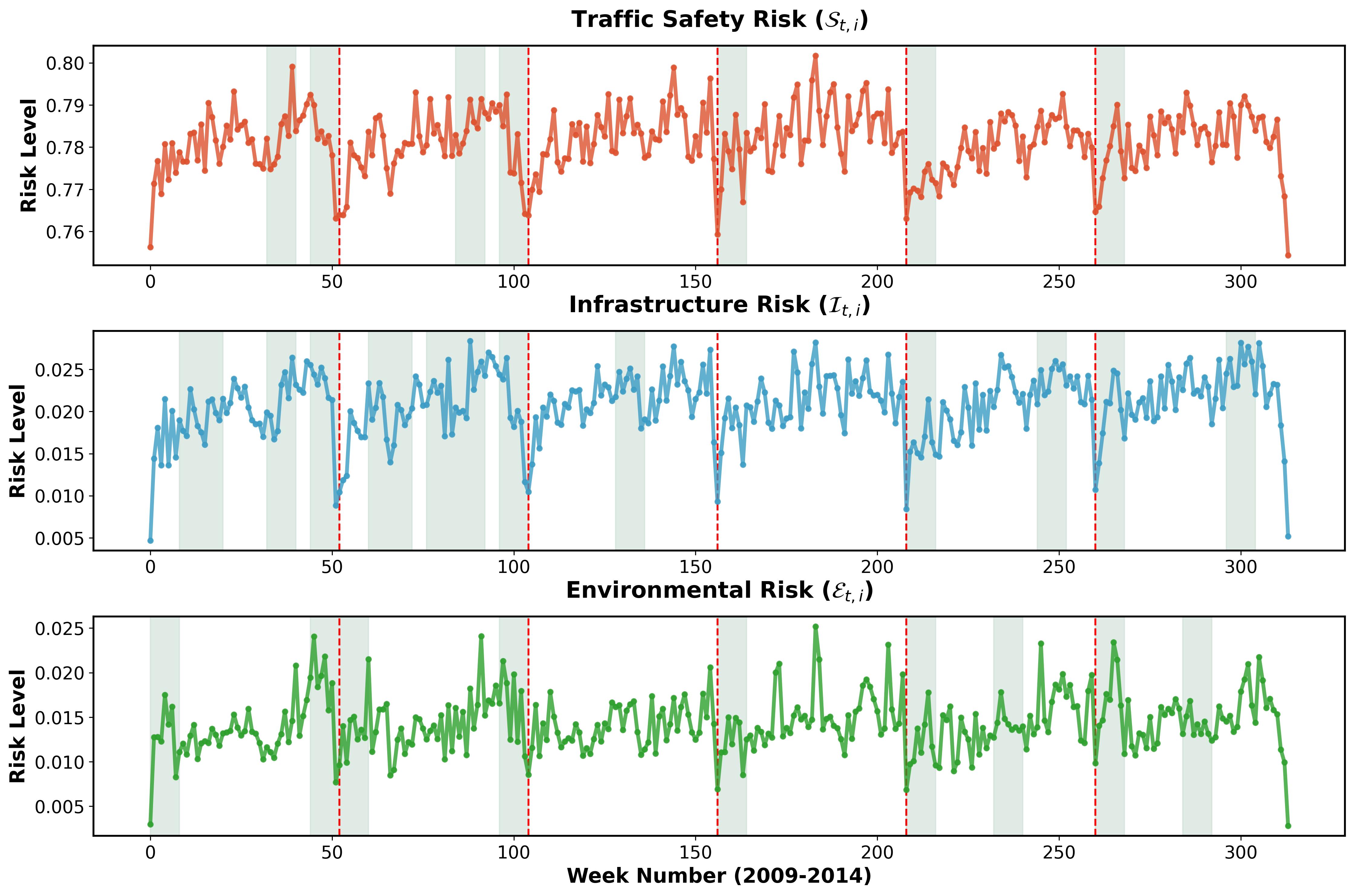}
    \captionof{figure}{Temporal risk evolution for three features}
    \label{fig:three_feature}
\end{center}

\subsection{Spatial Diffusion Modeling}

The fundamental premise for spatial diffusion rests on Tobler's First Law of Geography, which states that "everything is related to everything else, but near things are more related than distant things" (\cite{Tobler:1970a}). In the context of road safety, Road systems constitute connected networks where topological relationships create non-independent observation units (\cite{Wang:2006a}). Accident risk demonstrates spatial dependence characteristics where hazardous conditions on one road segment influence adjacent segments (\cite{Xu:2015a}). 

Under the theoretical background, spatial diffusion represents a critical methodological component in our road accident risk modeling framework, providing a mechanism to account for the complex inter-dependencies between proximal road segments. First, a sparse adjacency matrix  $\mathbf{A} \in \mathbb{R}^{N \times N}$ is constructed using Gaussian kernels to capture proximity-based relationships while limiting connections to k-nearest neighbors, ensuring that spatially closer road segments have stronger influence weights:

$$A_{i,j} = \begin{cases}
\exp\left(-\frac{d_{ij}^2}{2\sigma^2}\right) & \text{if } j \in \mathcal{N}_k(i) \\
0 & \text{otherwise}
\end{cases}$$

where $d_{ij}$ represents the Haversine geographical distance between nodes $i$ and $j$:

$$d_{ij} = 2R \arcsin\left(\sqrt{\sin^2\left(\frac{\phi_j - \phi_i}{2}\right) + \cos(\phi_i)\cos(\phi_j)\sin^2\left(\frac{\lambda_j - \lambda_i}{2}\right)}\right)$$

Here, $R$ is the Earth's radius, and $(\phi_i, \lambda_i)$ and $(\phi_j, \lambda_j)$ represent the latitude and longitude (in radians) of nodes $i$ and $j$, respectively. $\mathcal{N}_k(i)$ denotes the $k$-nearest neighbor set of node $i$, and $\sigma$ is the bandwidth parameter of the Gaussian kernel function.

Next, to ensure that the spectral properties of the matrix remain stable, which is essential for convergent iterative diffusion operations, the method applies symmetric normalization to the adjacency matrix:

$$\widetilde{\mathbf{A}} = \mathbf{D}^{-\frac{1}{2}}\mathbf{A}\mathbf{D}^{-\frac{1}{2}}$$

where $\mathbf{D}$ is the degree matrix with $D_{ii} = \sum_{j=1}^{N} A_{ij}$.

Furthermore, to address the differences in spatial propagation characteristics across feature dimensions, this framework designs a differentiated diffusion strategy. For time step $t$ and feature dimension $f$, the diffusion process is defined as:

$$\mathbf{X}^{(l+1)}_{t,:,f} = (1-\alpha_f)\mathbf{X}^{(l)}_{t,:,f} + \alpha_f \widetilde{\mathbf{A}}\mathbf{X}^{(l)}_{t,:,f}$$

where $l$ represents the diffusion iteration number, and $\alpha_f$ is the feature-specific diffusion coefficient that captures the varying degrees of spatial autocorrelation inherent to different risk dimensions. The optimal values of $\alpha_f$ and $l$ are determined through systematic ablation studies to reflect the distinct spatial propagation behaviors of accident severity, infrastructure characteristics, and environmental conditions.

Lastly, to preserve the authority of original observational data, the framework applies a weighted fusion mechanism after each diffusion step:

$$\mathbf{X}^{(\text{final})}_{t,i,f} = \beta \cdot \mathbf{X}^{(\text{diffused})}_{t,i,f} + (1-\beta) \cdot \mathbf{X}^{(\text{original})}_{t,i,f}$$

where $\beta = 0.7$ is the fusion weight, ensuring that the diffusion results can capture spatial dependencies while maintaining the local specificity of original data.

The final spatiotemporal coupled representation is obtained through multi-scale spatiotemporal fusion:

$$\mathcal{H}_{t,i} = \text{Concat}\left(\mathbf{X}^{(\text{final})}_{t,i,0}, \mathbf{X}^{(\text{final})}_{t,i,1}, \mathbf{X}^{(\text{final})}_{t,i,2}\right)$$

Through this multi-level spatiotemporal coupling and differentiated spatial diffusion mechanism, the framework can effectively model the complex spatiotemporal dynamics of urban traffic accidents, providing feature representations rich in spatiotemporal semantic information for predictive models. The resultant spatial distribution patterns, as demonstrated in Fig.\ref{fig:spatial_distribution}, reveal distinct geographical clustering characteristics across the three risk dimensions, thereby validating the theoretical foundation for spatially-aware traffic safety modeling.

\begin{center}
    \includegraphics[width=\textwidth]{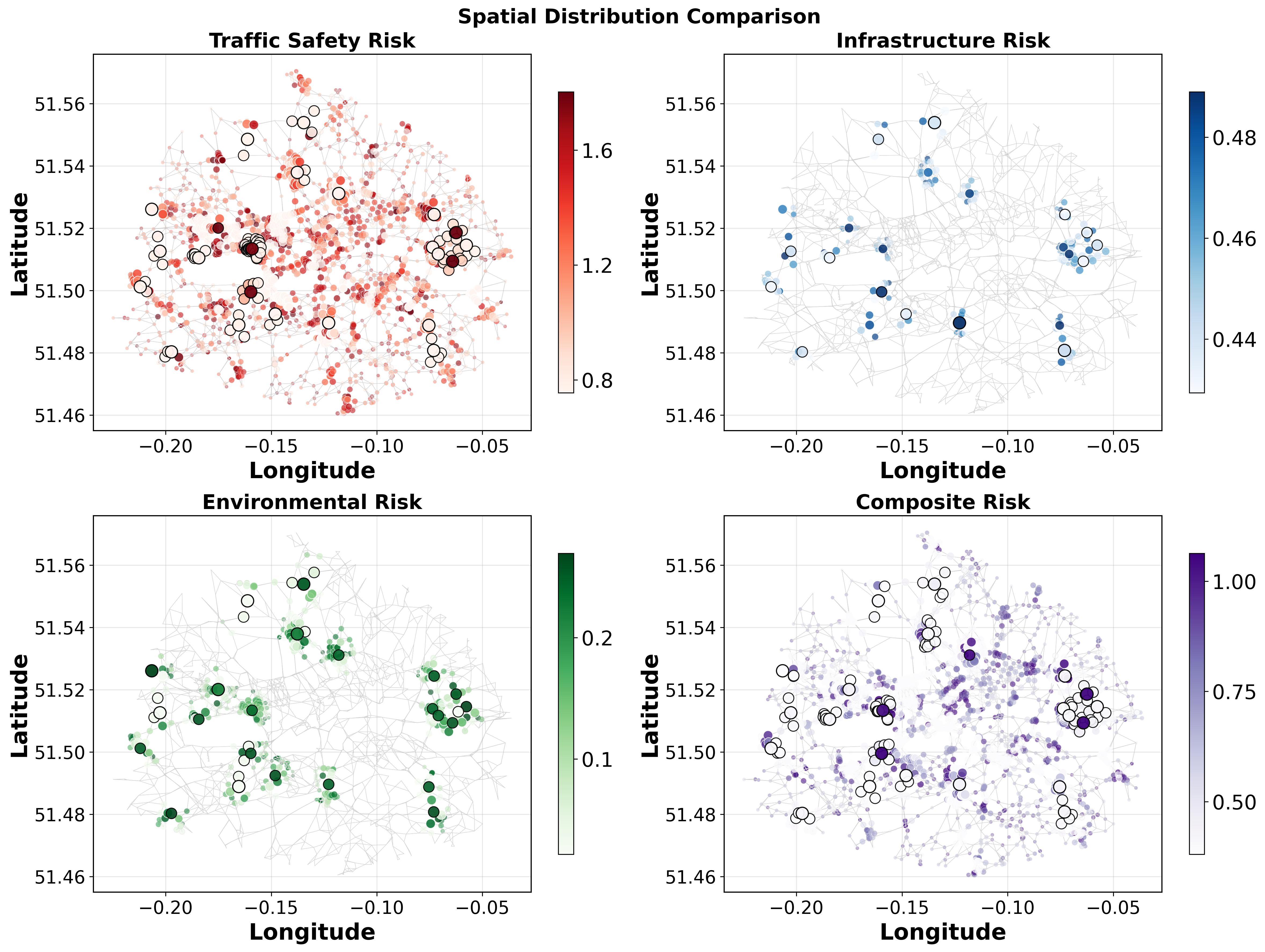}
    \captionof{figure}{Spatial distribution for three features after iterated diffusion}
    \label{fig:spatial_distribution}
\end{center}

\subsection{Spatio-Temporal Graph Neural Network Architecture}

This study proposes a Multi-Dimensional Attention Spatio-Temporal Graph Neural Network (MDAS-GNN) which extends established STGNN principles with multi-dimensional risk modeling for traffic safety applications (as shown in Fig.\ref{fig:diagram_architecture}). The framework combines feature-specific spatial attention mechanisms with multi-head temporal attention. This captures both spatial propagation patterns and long-range temporal dependencies, enabling selective information processing across different risk dimensions and time horizons.

\begin{center}
    \includegraphics[width=\textwidth]{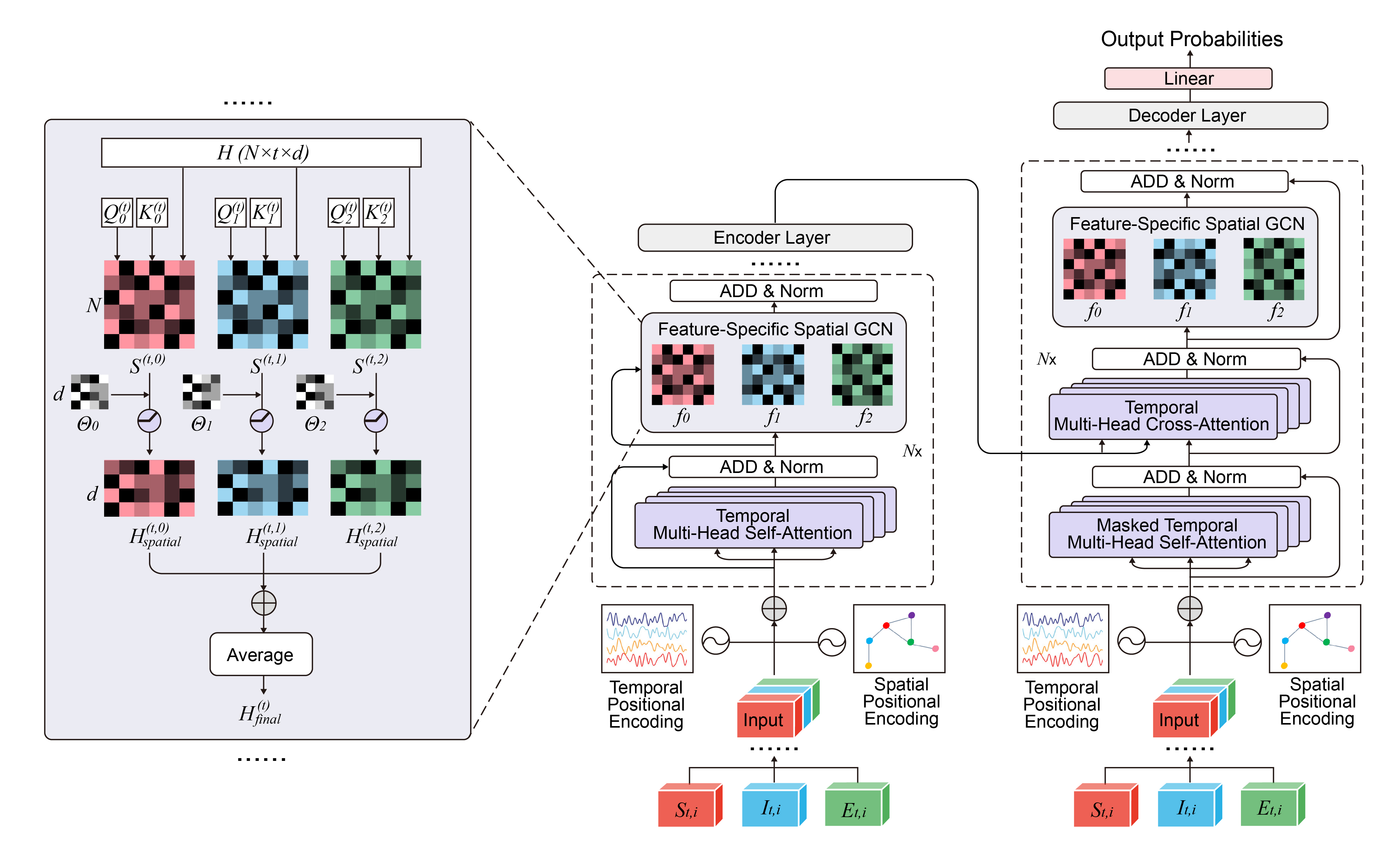}
    \captionof{figure}{MDAS-GNN: Model architecture}
    \label{fig:diagram_architecture}
\end{center}

\subsubsection{Feature-Specific Spatial Graph Convolution Module}

The feature-specific spatial graph convolution module is specifically designed to process multi-dimensional embeddings. Given the embedded feature matrix $\mathbf{H} \in \mathbb{R}^{N \times T \times d}$ (where the original three-dimensional risk features $\mathbf{X}^{(\text{input})} \in \mathbb{R}^{N \times T \times 3}$ have been projected to $d$-dimensional embeddings), the spatial module aggregates risk information across connected road segments to capture the propagation patterns of safety hazards.

The core spatial processing operates on the normalized adjacency matrix $\widetilde{\mathbf{A}}$ with standard graph convolution:

\begin{equation}
\mathbf{H}^{(l+1)}_{t,:,:} = \sigma\left(\widetilde{\mathbf{A}} \mathbf{H}^{(l)}_{t,:,:} \mathbf{\Theta}^{(l)}\right)
\end{equation}

where $\mathbf{H}^{(l)}_{t,:,:} \in \mathbb{R}^{N \times d}$ represents the embedded features at time step $t$ and layer $l$. The transformation matrix $\mathbf{\Theta}^{(l)} \in \mathbb{R}^{d \times d}$ enables the network to discover complex spatial patterns within the embedding space.

To address the heterogeneous nature of the original risk features, the framework implements feature-specific spatial attention that applies different attention patterns to the unified embedding space, recognizing that the three original risk types (traffic risk, infrastructure, and environmental) require distinct spatial propagation behaviors:

\begin{equation}
\mathbf{S}^{(t,f)} = \text{softmax}\left(\frac{\mathbf{Q}_f^{(t)} (\mathbf{K}_f^{(t)})^T}{\sqrt{d}}\right)
\end{equation}

\begin{equation}
\mathbf{H}_{\text{spatial}}^{(t,f)} = \sigma\left(\mathbf{\Theta}_f \left(\widetilde{\mathbf{A}} \odot \mathbf{S}^{(t,f)}\right) \mathbf{H}^{(t)}\right)
\end{equation}

where $f \in \{0,1,2\}$ indexes the three feature-specific attention patterns, $\mathbf{Q}_f^{(t)}, \mathbf{K}_f^{(t)} = \mathbf{W}_f^{(Q,K)} \mathbf{H}^{(t)} \in \mathbb{R}^{N \times d}$ are feature-specific query and key projections of the unified embeddings, and $\mathbf{\Theta}_f \in \mathbb{R}^{d \times d}$ represents feature-specific transformation matrices. The final output combines the three attended representations:

\begin{equation}
\mathbf{H}_{\text{final}}^{(t)} = \frac{1}{3}\sum_{f=0}^{2} \mathbf{H}_{\text{spatial}}^{(t,f)}
\end{equation}

This design acknowledges that traffic risk patterns (pattern 0) require stronger local spatial clustering due to traffic flow patterns, infrastructure risks (pattern 1) need moderate spatial correlation due to traffic control systems and faciliteis, and environmental risks (pattern 2) benefit from regional correlation patterns due to weather and surface conditions affecting broader areas.

\vspace{\baselineskip}
\subsubsection{Multi-Head Temporal Attention Mechanism}

The temporal attention mechanism is specifically designed to handle the weekly-scale temporal patterns. Due to the complex temporal dependencies of weekly traffic safety patterns, traditional recurrent architectures are unable to fully capture these dependencies due to their sequential processing nature and vanishing gradient problems (\cite{Bengio:1994a}).

The approach implements weekly-focused temporal attention that addresses the specific characteristics of accident prediction at this temporal scale. The temporal attention mechanism processes weekly sequences through 1D convolutions that capture local temporal context within the weekly time series. The input weekly aggregated embedded features $\mathbf{H}_{\text{weekly}} \in \mathbb{R}^{N \times T_{weeks} \times d}$ are transformed into query, key, and value representations through temporal convolutions and linear projections:

\begin{equation}
\mathbf{Q}^{(h)}, \mathbf{K}^{(h)}, \mathbf{V}^{(h)} = f_{\text{conv}}(\mathbf{H}_{\text{weekly}}) \mathbf{W}^{(h)}
\end{equation}

where $f_{\text{conv}}(\cdot)$ represents 1D convolution operations with kernel sizes optimized for weekly pattern detection, and $\mathbf{W}^{(h)} \in \mathbb{R}^{d \times d}$ denotes the learnable projection matrices for attention head $h$. The convolution operations enable the attention mechanism to incorporate temporal context from adjacent weeks and capture temporal dependencies in accident patterns.

Based on these temporal convolution operations, causal masking is applied to ensure proper autoregressive behavior. For autoregressive weekly prediction, causal attention mechanisms ensure that predictions only depend on past and current weekly observations through masking:

\begin{equation}
\text{Attention}_{\text{causal}}(\mathbf{Q}, \mathbf{K}, \mathbf{V}) = \text{softmax}\left(\frac{\mathbf{Q}\mathbf{K}^T}{\sqrt{d_k}} + \text{Mask}\right) \mathbf{V}
\end{equation}

where the mask prevents future information leakage by setting attention weights to zero for future time steps. This weekly-focused approach aligns with the temporal resolution that provides optimal signal-to-noise ratio for accident prediction.

\vspace{\baselineskip}
\subsubsection{Encoder-Decoder Framework}

The encoder-decoder architecture compresses input sequences into a fixed representation, then expands this back into output sequences. The encoder distills patterns from the entire input into a compact vector, while the decoder uses this representation to generate outputs step-by-step (\cite{Sutskever:2014a}). Each encoder layer implements a residual connection around two sub-layers: temporal multi-head attention followed by feature-specific spatial graph convolution. The layer design follows the pre-normalization structure for improved training stability (\cite{Xiong:2020a}):

\begin{equation}
\mathbf{H}_{\text{attn}}^{(l)} = \mathbf{H}^{(l-1)} + \text{TemporalAttn}(\text{LayerNorm}(\mathbf{H}^{(l-1)}))
\end{equation}

\begin{equation}
\mathbf{H}_{\text{encoder}}^{(l)} = \mathbf{H}_{\text{attn}}^{(l)} + \text{FeatureSpecificSpatialGCN}(\text{LayerNorm}(\mathbf{H}_{\text{attn}}^{(l)}))
\end{equation}

where the residual connections facilitate gradient flow through deep networks, and layer normalization (\cite{Ba:2016a}) stabilizes training by normalizing inputs to each sub-layer. The sequential application of temporal attention followed by feature-specific spatial convolution allows the model to first capture temporal patterns within each node's time series, then propagate and integrate this information across the spatial graph structure using distinct patterns for different risk types.

The decoder implements a three-step attention mechanism that progressively builds predictions using causal temporal self-attention, encoder-decoder cross-attention, and feature-specific spatial graph convolution:

\begin{equation}
\mathbf{H}_{\text{self}}^{(l)} = \mathbf{H}_{\text{dec}}^{(l-1)} + \text{MaskedTemporalAttn}(\text{LayerNorm}(\mathbf{H}_{\text{dec}}^{(l-1)}))
\end{equation}

\begin{equation}
\mathbf{H}_{\text{cross}}^{(l)} = \mathbf{H}_{\text{self}}^{(l)} + \text{CrossAttn}(\text{LayerNorm}(\mathbf{H}_{\text{self}}^{(l)}), \mathbf{H}_{\text{encoder}})
\end{equation}

\begin{equation}
\mathbf{H}_{\text{decoder}}^{(l)} = \mathbf{H}_{\text{cross}}^{(l)} + \text{FeatureSpecificSpatialGCN}(\text{LayerNorm}(\mathbf{H}_{\text{cross}}^{(l)}))
\end{equation}

The masked temporal self-attention ensures causal dependencies in the output sequence, while the cross-attention mechanism allows the decoder to selectively attend to relevant parts of the encoded input sequence. The final feature-specific spatial graph convolution integrates spatial information into the prediction process.

The final prediction layer transforms the decoder output to accident risk probabilities through a linear transformation. The linear transformation maps the $d$-dimensional hidden representations to single-dimensional risk scores for each road segment, providing interpretable predictions that can be used for safety analysis and intervention planning.

\section{Experiment}

\subsection{Experimental Setup}

The data is temporally partitioned into training (60\%), validation (20\%), and test (20\%) sets. The model employs a single-layer architecture with spatial embedding and temporal encoding mechanisms. Training is conducted over 50 primary epochs followed by 20 fine-tuning epochs using the Adam optimizer with an initial learning rate of $1 \times 10^{-5}$, reduced by a factor of 10 during fine-tuning. The experimental setup utilizes L1 loss with model checkpointing based on validation performance.
Model performance is assessed using three standard regression metrics. Let $\hat{y}_{i,t}$ and $y_{i,t}$ denote the predicted and ground truth values for node $i$ at time step $t$, respectively.

\textbf{Mean Absolute Error (MAE)} quantifies the average absolute deviation:
\begin{equation}
\text{MAE} = \frac{1}{N \times T} \sum_{i=1}^{N} \sum_{t=1}^{T} |\hat{y}_{i,t} - y_{i,t}|
\end{equation}

\textbf{Root Mean Square Error (RMSE)} measures the standard deviation of prediction residuals:
\begin{equation}
\text{RMSE} = \sqrt{\frac{1}{N \times T} \sum_{i=1}^{N} \sum_{t=1}^{T} (\hat{y}_{i,t} - y_{i,t})^2}
\end{equation}

\textbf{Mean Absolute Percentage Error (MAPE)} provides scale-independent evaluation:
\begin{equation}
\text{MAPE} = \frac{100\%}{N \times T} \sum_{i=1}^{N} \sum_{t=1}^{T} \frac{|\hat{y}_{i,t} - y_{i,t}|}{|y_{i,t}|}
\end{equation}

MAPE serves as the primary evaluation criterion due to its scale-independence, enabling fair comparison across road segments with differing baseline accident rates.

\subsection{Experiment Results}

This study evaluates the performance for traffic safety prediction across three distinct temporal horizons: short-term (1-4 weeks), medium-term (5-8 weeks), and long-term (9-12 weeks). While the main methodology focused on the London case study, the experimental validation was extended to encompass three metropolitan regions in the UK, including Center London, South Manchester, and SE Birmingham, to assess the generalizability and robustness of the proposed approach across different urban contexts.

The experimental comparison includes six established baseline models alongside the proposed MDAS-GNN: STGNN (2020), ASTGCN (2019), STGCN (2018), Graph WaveNet (2019), LSTM (1997), and SVR (1995). These baselines represent diverse approaches to spatiotemporal prediction, ranging from traditional machine learning methods (SVR, LSTM) to state-of-the-art graph-based neural architectures (STGCN, ASTGCN, Graph WaveNet, STGNN).

The proposed MDAS-GNN model demonstrates exceptional performance in short-term predictions, achieving the lowest error rates across most metrics and regions (as shown in Table.\ref{table:short_term_performance}). Specifically, it attains MAPE values of 3.707\% (Center London), 3.183\% (South Manchester), and 3.027\% (SE Birmingham) for 1-4 week forecasts. The model consistently outperforms traditional approaches such as LSTM (21.080 \%, 10.863\%, 6.460\%) and SVR (14.313\%, 10.693\%, 29.050\%). Among graph-based methods, STGCN emerges as the strongest competitor, while ASTGCN shows moderate performance across regions.

Medium-term predictions reveal interesting performance dynamics (as shown in Table.\ref{table:medium_term_performance}), with STGCN achieving optimal results in Center London (2.863\% MAPE) and South Manchester (2.363\% MAPE). MDAS-GNN maintains competitive performance with MAPE values of 3.480\%, 3.017\%, and 2.973\% across the three regions, while ASTGCN shows regional variability with significant degradation in South Manchester (19.447\%).

Long-term predictions showcase MDAS-GNN's strength (as shown in Table.\ref{table:long_term_performance}), maintaining superior performance with MAPE values of 4.343\%, 2.993\%, and 3.880\% for 9--12 week forecasts. STGCN continues competitive performance (6.360\%, 2.568\%, 3.623\%), while ASTGCN exhibits severe degradation, particularly in Center London (42.148\% MAPE).

The detailed weekly performance analysis reveals significant variations in temporal stability across models (as shown in Fig.\ref{fig:stability_comparison}). MDAS-GNN exhibits remarkable consistency with MAPE fluctuations within narrow ranges: 3.29--3.96\% for Center London, 2.79--3.6\% for South Manchester, and 2.91--3.71\% for SE Birmingham. This contrasts sharply with competing models, where STGNN shows extreme volatility (5.49\% to 52.33\% in Center London) and ASTGCN demonstrates progressive degradation (3.74\% to 68.13\% MAPE in Center London).

The experimental results demonstrate that MDAS-GNN achieves superior performance and temporal stability compared to baseline methods across three UK metropolitan regions. The consistent 2--3\% MAPE performance across 12-week prediction periods validates that spatiotemporal graph neural networks can effectively capture seasonal traffic safety patterns while filtering stochastic noise, representing significant advancements for traffic safety management systems requiring long-term planning decisions.

\newpage
\begin{center}
\captionof{table}{Short-term prediction performance comparison (1-4 weeks)}
\label{table:short_term_performance}
\footnotesize
\begin{tabular}{l c c c c c c c c c}
\hline\hline
\multirow{2}{*}{Model} & 
\multicolumn{3}{c}{Center London} & 
\multicolumn{3}{c}{South Manchester} & 
\multicolumn{3}{c}{SE Birmingham} \\
\cline{2-10}
& MAE & RMSE & MAPE & MAE & RMSE & MAPE & MAE & RMSE & MAPE \\
& (1) & (2) & (3) & (4) & (5) & (6) & (7) & (8) & (9) \\
\hline
\textbf{MDAS-GNN} & \textbf{0.043} & \textbf{0.190} & \textbf{3.707} & \textbf{0.040} & \textbf{0.190} & \underline{3.183} & \textbf{0.040} & \textbf{0.180} & \textbf{3.027} \\
STGNN & 0.150 & 0.243 & 17.593 & 0.110 & 0.223 & 12.607 & 0.267 & 0.340 & 16.81 \\
ASTGCN & 0.097 & 0.210 & 10.577 & \underline{0.047} & \textbf{0.190} & 3.633 & \textbf{0.040} & \underline{0.183} & \underline{3.493} \\
STGCN & \underline{0.047} & \underline{0.193} & \underline{4.117} & 0.043 & \underline{0.193} & \textbf{3.150} & \underline{0.047} & \underline{0.183} & 4.243 \\
GraphWave & 0.083 & 0.203 & 8.893 & 0.067 & \underline{0.193} & 6.560 & 0.133 & 0.213 & 15.473 \\
LSTM & 0.627 & 0.683 & 21.080 & 0.133 & 0.237 & 10.863 & 0.063 & 0.187 & 6.460 \\
SVR & 0.123 & 0.243 & 14.313 & 0.097 & 0.207 & 10.693 & 0.233 & 0.300 & 29.050 \\
\hline\hline
\end{tabular}
\normalsize

\vspace{0.2em}
\parbox{\linewidth}{\footnotesize Note: \textbf{Bold} numbers indicate the best performance, \underline{Underlined numbers} indicate the second best.}
\end{center}

\begin{center}
\captionof{table}{Medium-term prediction performance comparison (5-8 weeks)}
\label{table:medium_term_performance}
\footnotesize
\begin{tabular}{l c c c c c c c c c}
\hline\hline
\multirow{2}{*}{Model} & 
\multicolumn{3}{c}{Center London} & 
\multicolumn{3}{c}{South Manchester} & 
\multicolumn{3}{c}{SE Birmingham} \\
\cline{2-10}
& MAE & RMSE & MAPE & MAE & RMSE & MAPE & MAE & RMSE & MAPE \\
& (1) & (2) & (3) & (4) & (5) & (6) & (7) & (8) & (9) \\
\hline
\textbf{MDAS-GNN} & \underline{0.047} & \textbf{0.190} & \underline{3.480} & \underline{0.040} & \textbf{0.190} & \underline{3.017} & \textbf{0.037} & \textbf{0.180} & \underline{2.973} \\
STGNN & 0.070 & 0.193 & 7.010 & 0.227 & 0.290 & 28.343 & 0.233 & 0.307 & 29.407 \\
ASTGCN & 0.053 & 0.200 & 4.793 & 0.167 & 0.253 & 19.447 & 0.050 & \underline{0.183} & 4.640 \\
STGCN & \textbf{0.040} & \textbf{0.190} & \textbf{2.863} & \textbf{0.030} & \underline{0.193} & \textbf{2.363} & \underline{0.040} & \textbf{0.180} & \textbf{2.833} \\
GraphWave & 0.067 & 0.193 & 6.413 & \underline{0.040} & \textbf{0.190} & 3.393 & 0.180 & 0.267 & 21.690 \\
LSTM & \underline{0.047} & \underline{0.191} & 4.000 & 0.043 & \textbf{0.190} & 3.167 & \textbf{0.037} & \textbf{0.180} & 2.987 \\
SVR & 0.123 & 0.230 & 14.210 & 0.090 & 0.203 & 9.887 & 0.230 & 0.277 & 28.947 \\
\hline\hline
\end{tabular}
\normalsize

\vspace{0.2em}
\parbox{\linewidth}{\footnotesize Note: \textbf{Bold} numbers indicate the best performance, \underline{Underlined numbers} indicate the second best.}
\end{center}

\begin{center}
\captionof{table}{Long-term prediction performance comparison (9--12 weeks)}
\label{table:long_term_performance}
\footnotesize
\begin{tabular}{l c c c c c c c c c}
\hline\hline
\multirow{2}{*}{Model} & 
\multicolumn{3}{c}{Center London} & 
\multicolumn{3}{c}{South Manchester} & 
\multicolumn{3}{c}{SE Birmingham} \\
\cline{2-10}
& MAE & RMSE & MAPE & MAE & RMSE & MAPE & MAE & RMSE & MAPE \\
& (1) & (2) & (3) & (4) & (5) & (6) & (7) & (8) & (9) \\
\hline
\textbf{MDAS-GNN} & \textbf{0.053} & \textbf{0.190} & \textbf{4.343} & \underline{0.040} & \textbf{0.190} & \underline{2.993} & 0.045 & \textbf{0.180} & 3.880 \\
STGNN      & 0.083 & 0.203 & 8.745 & 0.120 & 0.238 & 13.893 & 0.430 & 0.473 & 30.340 \\
ASTGCN     & 0.328 & 0.363 & 42.148 & 0.123 & 0.218 & 14.468 & \underline{0.040} & \underline{0.183} & \underline{3.155} \\
STGCN      & \underline{0.063} & \underline{0.193} & 6.360 & \textbf{0.038} & \underline{0.193} & \textbf{2.568} & 0.045 & \underline{0.183} & 3.623 \\
GraphWave  & 0.175 & 0.248 & 21.330 & 0.050 & \textbf{0.190} & 4.685 & 0.153 & 0.240 & 18.578 \\
LSTM       & \textbf{0.053} & \underline{0.193} & \underline{4.915} & 0.048 & \underline{0.193} & 3.935 & \textbf{0.035} & \textbf{0.180} & \textbf{2.845} \\
SVR        & 0.443 & 0.460 & 57.758 & 0.113 & 0.210 & 12.613 & 0.393 & 0.428 & 49.940 \\
\hline\hline
\end{tabular}
\normalsize

\vspace{0.2em}
\parbox{\linewidth}{\footnotesize Note: \textbf{Bold} numbers indicate the best performance, \underline{Underlined numbers} indicate the second best.}
\end{center}

\begin{center}
    \includegraphics[width=\textwidth]{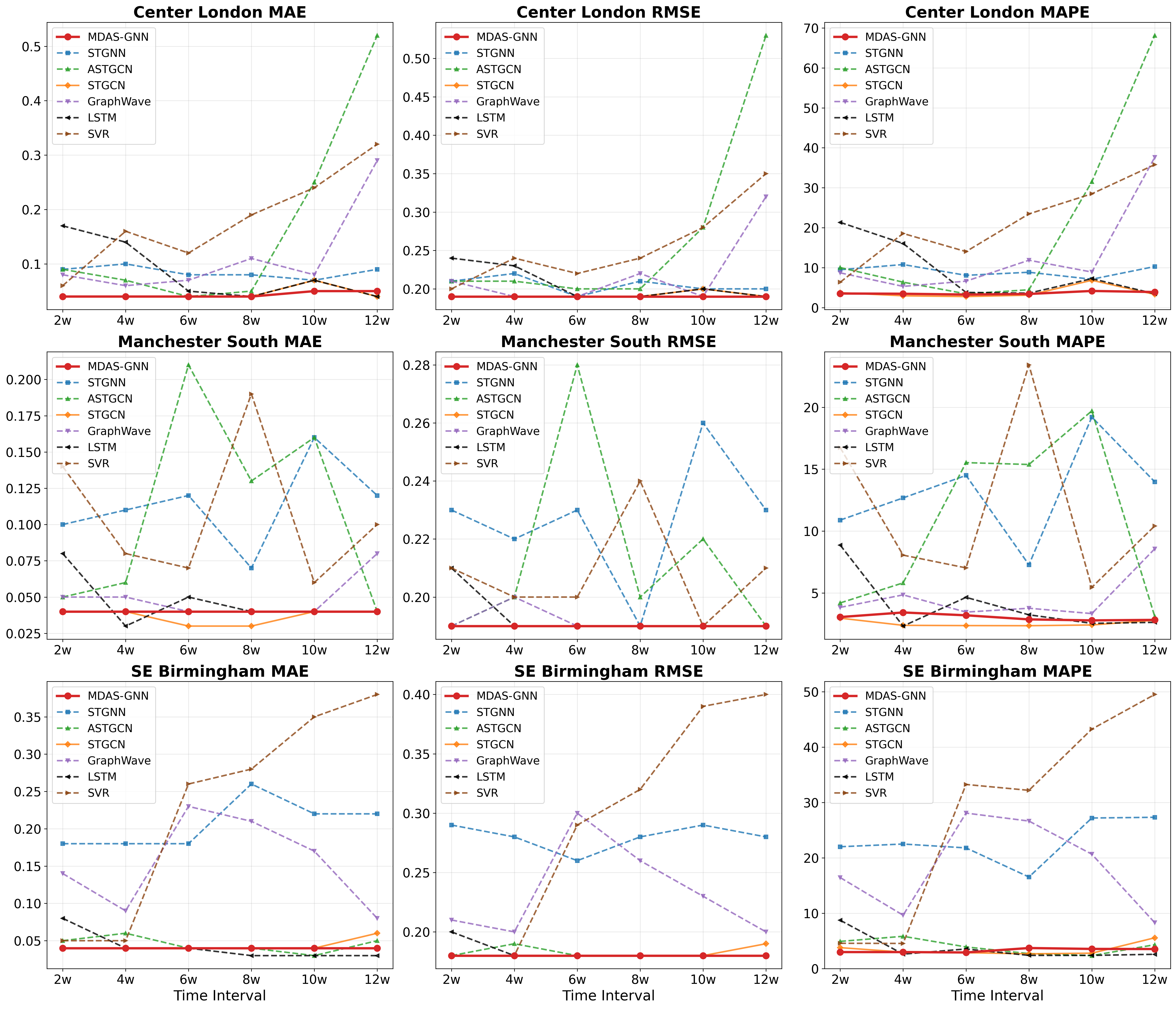}
    \captionof{figure}{Temporal stability comparison}
    \label{fig:stability_comparison}
\end{center}

\subsection{Ablation Study}

\subsubsection{Feature-Specific Ablation Analysis}

Understanding the individual contribution of each feature category is crucial for traffic risk forecasting models, as it helps identify the relative importance of traffic safety risk, infrastructure risk, and environmental risk features. This ablation study (as shown in Fig.\ref{fig:feature_ablation}) examines how different input feature combinations affect MDAS-GNN model performance across temporal horizons. Four configurations were tested: all features (traffic safety risk, environmental risk, infrastructure risk), traffic safety risk with environmental risk, traffic safety risk with infrastructure risk, and traffic risk only.

The complete feature set achieved optimal performance with the lowest mean absolute percentage error across all time-frames (3.070\% short-term, 3.215\% medium-term, 2.768\% long-term). This demonstrates that combining environmental and infrastructure data with key traffic risk indicators produces synergistic improvements in predictive accuracy. Environmental factors showed slightly greater influence than infrastructure features. The traffic-risk-only configuration yielded the poorest results, reaching 6.423\% MAPE for long-term forecasts, confirming that traffic-risk data alone is insufficient for robust temporal prediction.

These findings establish that comprehensive feature integration is essential for optimal MDAS-GNN performance, with environmental context providing marginally stronger predictive value than infrastructure characteristics when combined with traffic risk indicators.

\vspace{\baselineskip}
\begin{center}
    \includegraphics[width=\textwidth]{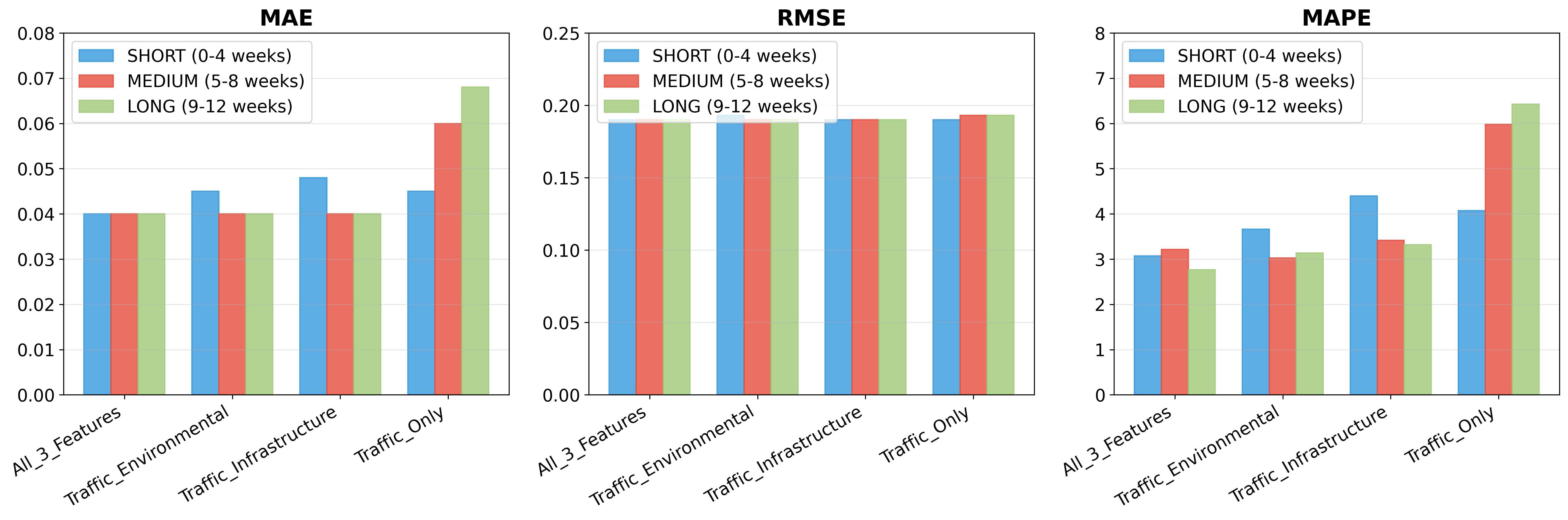}
    \captionof{figure}{Impact of feature combinations on MDAS-GNN performance across temporal horizons}
    \label{fig:feature_ablation}
\end{center}

\subsubsection{Spatial Diffusion Mechanism Analysis}

This ablation study systematically evaluates different risk diffusion parameter configurations to assess their impact on predictive model performance across temporal horizons. Eight experimental scenarios were tested, ranging from No\_Diffusion (baseline) to Over\_Diffusion, with varying intensities of traffic risk ($\alpha_0$), infrastructure risk ($\alpha_1$), and environmental risk ($\alpha_2$) parameters, with detailed parameter settings in Table \ref{table:diffusion_parameters}.

\newpage
\begin{center}
\captionof{table}{Ablation study configurations on feature-specific diffusion parameters}
\label{table:diffusion_parameters}
\footnotesize
\begin{tabular}{l c c c c c c}
\hline\hline
\multirow{2}{*}{Experiment} & 
\multicolumn{2}{c}{Traffic SafetyRisk} & 
\multicolumn{2}{c}{Infrastructure Risk} & 
\multicolumn{2}{c}{Environmental Risk} \\
\cline{2-7}
& ($\alpha_0$) & Traf\_Iter & ($\alpha_1$) & Infr\_Iter & ($\alpha_2$) & Env\_Iter \\
& (1) & (2) & (3) & (4) & (5) & (6) \\
\hline
No\_Diffusion & 0 & 0 & 0 & 0 & 0 & 0 \\
Uniform\_Weak & 0.1 & 1 & 0.1 & 1 & 0.1 & 1 \\
Uniform\_Medium & 0.2 & 1 & 0.2 & 1 & 0.2 & 1 \\
Uniform\_Strong & 0.3 & 2 & 0.3 & 2 & 0.3 & 2 \\
Differentiated\_Current & 0.2 & 1 & 0.2 & 1 & 0.2 & 1 \\
Differentiated\_A & 0.3 & 2 & 0.1 & 1 & 0.25 & 2 \\
Differentiated\_B & 0.25 & 1 & 0.15 & 1 & 0.3 & 2 \\
Over\_Diffusion & 0.5 & 3 & 0.4 & 3 & 0.4 & 3 \\
\hline\hline
\end{tabular}
\normalsize
\end{center}

\vspace{\baselineskip}

\begin{center}
    \includegraphics[width=0.9\textwidth]{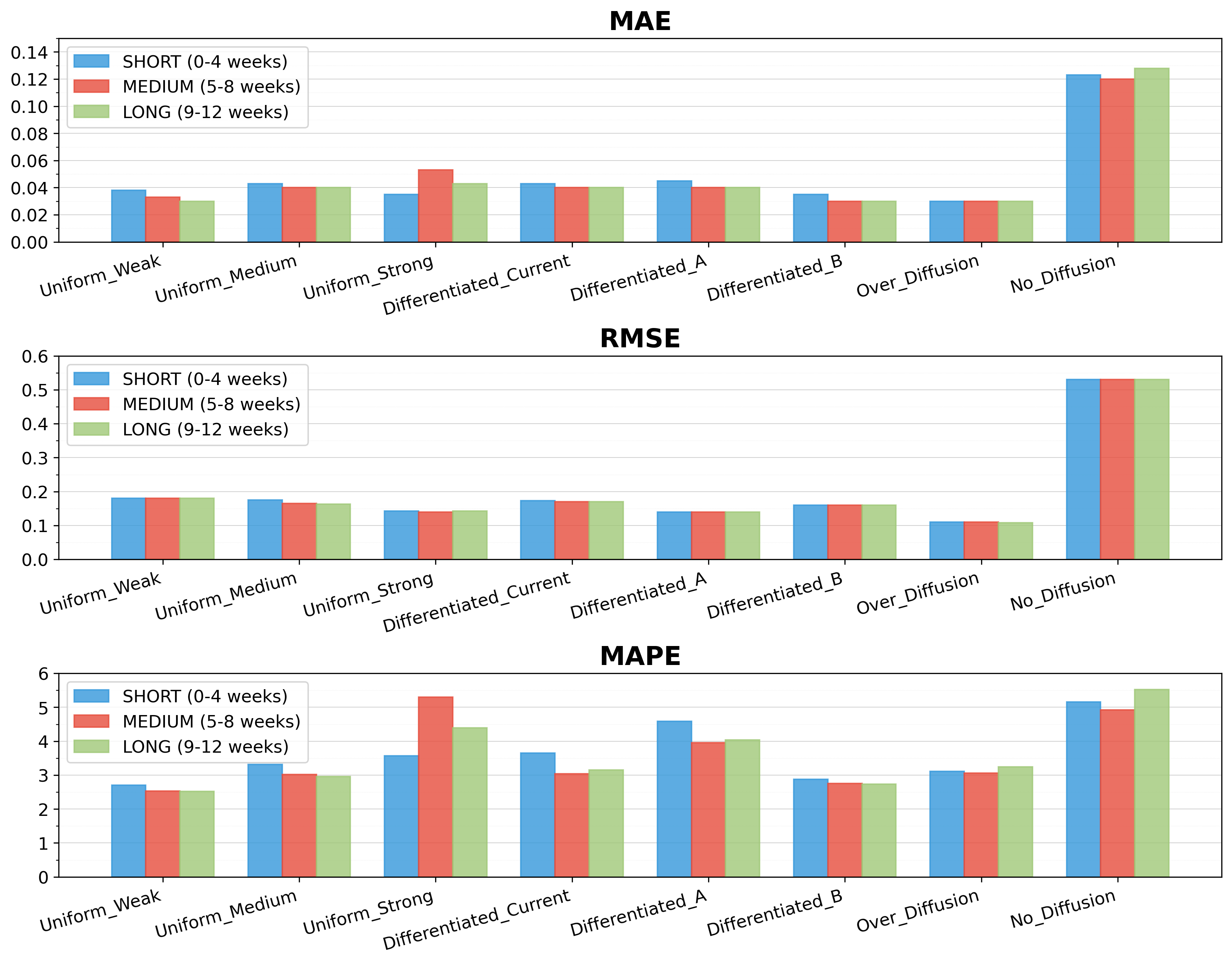}
    \captionof{figure}{Impact of feature-specific diffusion configurations on MDAS-GNN performance across temporal horizons}
    \label{fig:diffusion_ablation}
\end{center}

The Over\_Diffusion configuration achieved consistently strong performance with MAPE values of 3.11\% (short-term), 3.06\% (medium-term), and 3.24\% (long-term). Differentiated\_B performed exceptionally well, yielding the lowest overall errors (2.88\% short-term, 2.76\% medium-term, 2.73\% long-term MAPE), while No\_Diffusion produced substantially higher errors (5.16\% short-term, 4.92\% medium-term, 5.53\% long-term MAPE). 

Based on the comprehensive evaluation, the Differentiated\_B configuration ($\alpha_0 = 0.25$, $\alpha_1 = 0.15$, $\alpha_2 = 0.3$ with 1, 1, and 2 iterations respectively) emerges as the optimal setting (as shown in Fig.\ref{fig:diffusion_ablation}). These findings establish that balanced, differentiated risk parameter configurations outperform both under-diffused and uniformly aggressive diffusion strategies.

\section{Prediction and Engineering Applications}

Based on the MDAS-GNN framework's prediction outputs, this study generates comprehensive road traffic risk predictions that are systematically visualized through a sophisticated road-based mapping system. The framework predicts weekly accident occurrence probabilities across Central London's road network, with the model processing spatially-weighted prediction values and converting them into a percentile-based classification system, systematically ranking road segments into six distinct risk zones from no risk to very high risk across seasonal temporal horizons, as shown in Fig.\ref{fig:output_12steps}. Week 6 is enlarged in a detailed view to demonstrate the granular spatial resolution and risk differentiation capabilities, as shown in Fig.\ref{fig:time6}.

Civil engineers can leverage these visualized predictions for comprehensive safety planning and resource optimization across multiple operational scales. For immediate tactical decisions, engineers can use the detailed risk maps to prioritize emergency safety interventions, such as installing temporary traffic calming measures or enhanced signage at high-risk intersections, while environmental risk factors like adverse weather conditions  and poor road surface conditions can be addressed through real-time traffic management responses including speed restrictions and increased monitoring. For medium-term planning, the temporal evolution patterns enable strategic scheduling of infrastructure improvements, allowing engineers to coordinate major reconstruction projects with predicted high-risk periods and optimize construction timing to minimize safety impacts. For long-term strategic planning, the persistent high-risk corridors guide capital investment decisions for geometric redesign, intersection improvements, and systematic safety upgrades. Additionally, the spatial specificity of the visualizations enables engineers to differentiate between infrastructure-related risks requiring physical modifications and environmental risks needing operational responses, facilitating more targeted and cost-effective safety interventions that directly support Vision Zero objectives.

\begin{center}
    \includegraphics[width=\textwidth]{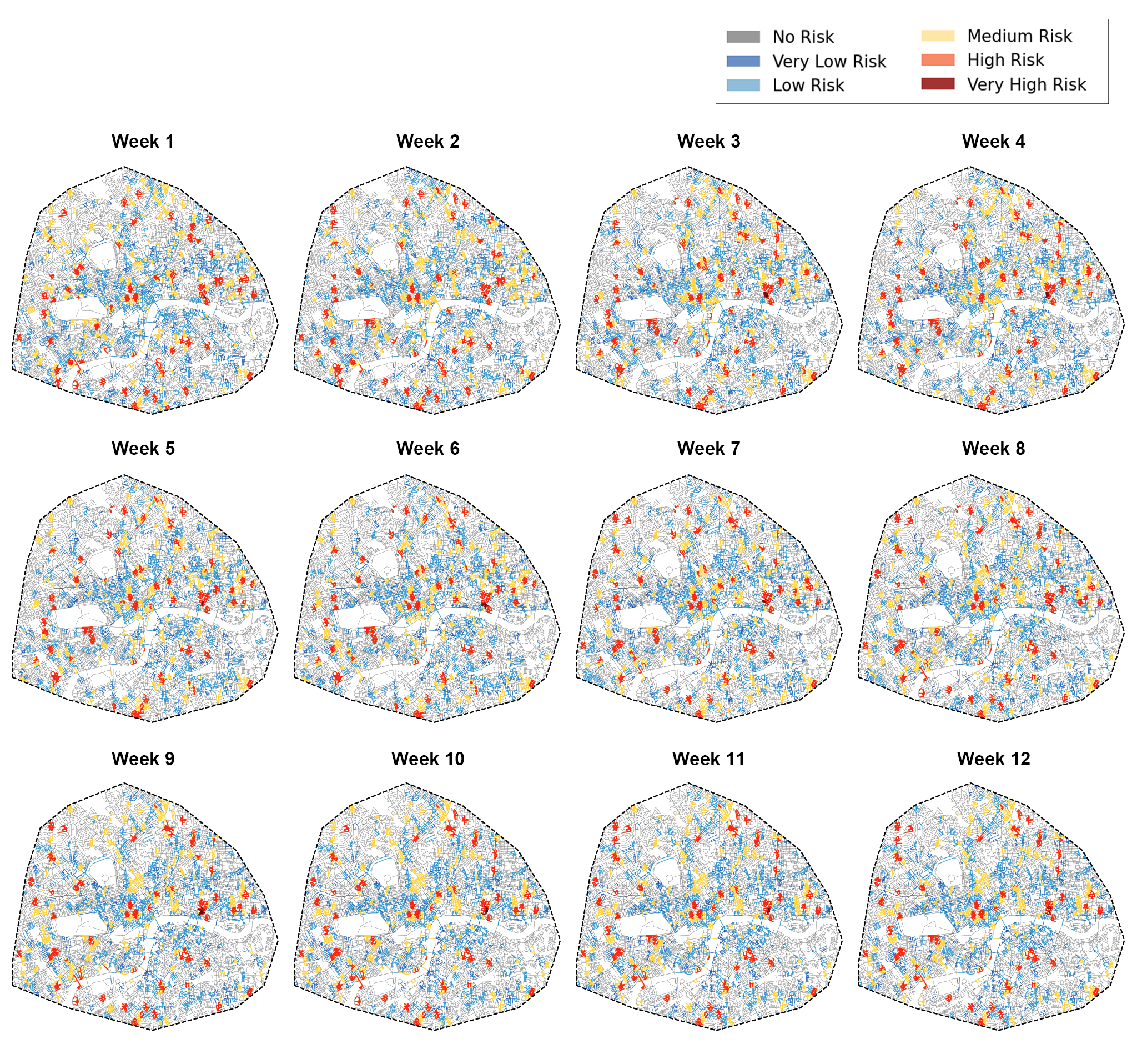}
    \captionof{figure}{12-week traffic safety risk prediction across central London road network}
    \label{fig:output_12steps}
\end{center}

\begin{center}
    \includegraphics[width=0.9\textwidth]{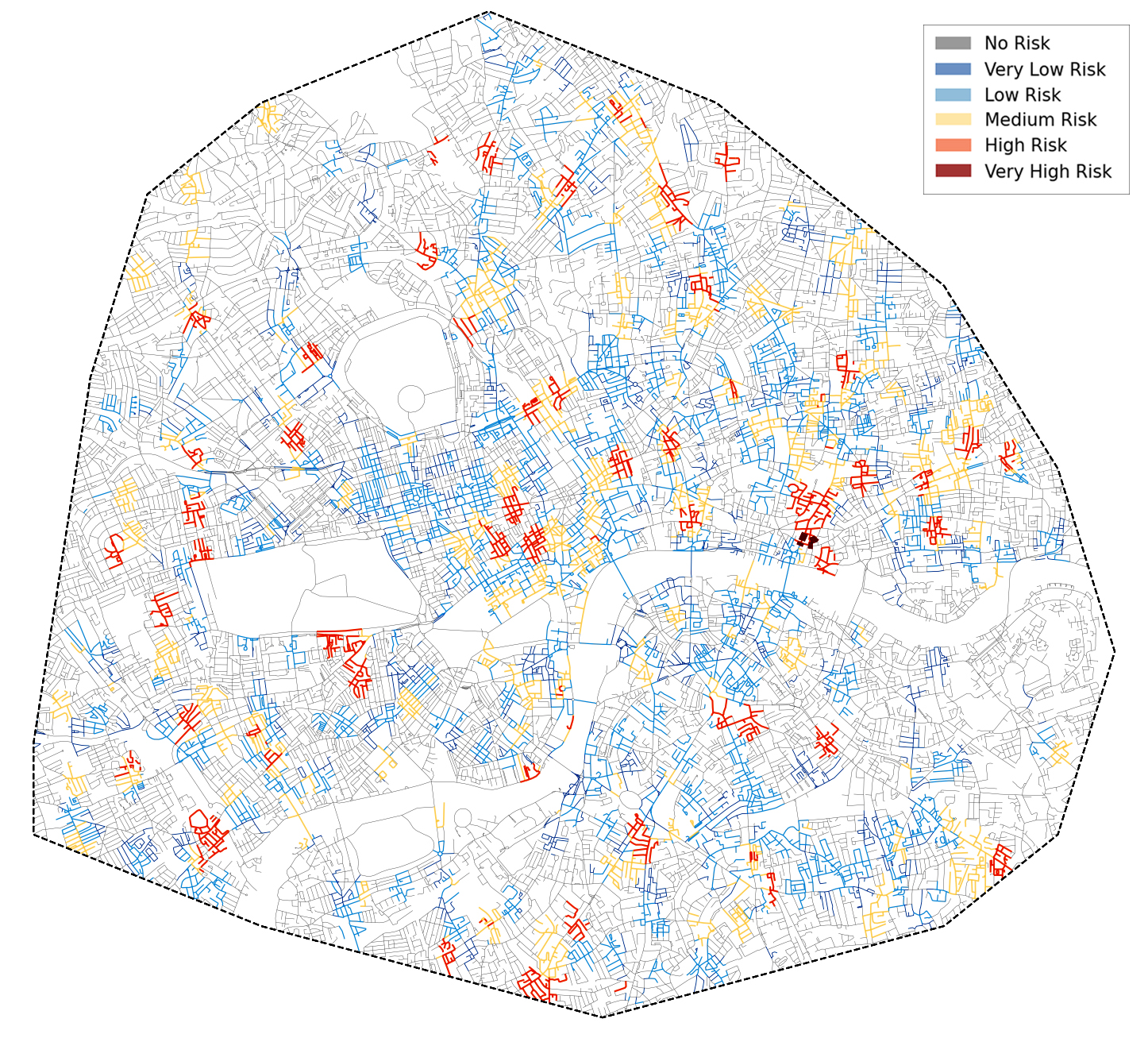}
    \captionof{figure}{Risk ranking map for week 6 with six-zone risk categorization}
    \label{fig:time6}
\end{center}

\section{Conclusion}

This research presents MDAS-GNN, an innovative spatiotemporal graph neural network that systematically integrates multi-dimensional risk factors for urban traffic accident prediction. The framework's feature-specific spatial diffusion mechanisms and temporal attention architecture successfully capture complex spatiotemporal dependencies that traditional models fail to address. Validation on comprehensive UK traffic datasets demonstrates substantial improvements in predictive accuracy and temporal consistency across multiple forecasting horizons. The model's ability to distinguish between different risk types enables targeted safety interventions and resource optimization. MDAS-GNN establishes a robust foundation for evidence-based transportation safety management, offering civil engineers and urban planners an advanced analytical tool for strategic infrastructure planning and accident prevention initiatives.

\section*{Declaration of Generative AI and AI-assisted technologies in the writing process}
During the preparation of this work the authors used generative AI to improve the English writing. After using this tool/service, the authors reviewed and edited the content as needed and take full responsibility for the content of the publication.

\section*{CRediT authorship contribution statement}
Ziyuan Gao: Conceptualization, Data curation, Formal analysis, Investigation, Methodology, Project administration, Resources, Software, Validation, Writing -- original draft, Writing -- review \& editing.

\section*{Declaration of competing interest}
The authors declare that they have no known competing financial interests or personal relationships that could have appeared to influence the work reported in this paper.

\section*{Funding}
This research did not receive any specific grant from funding agencies in the public, commercial, or not-for-profit sectors.

\section*{Data availability}
Data will be made available on request.


\begin{thebibliography}{00}

\bibitem[World Health Organization, 2018]{WHO:2018}
World Health Organization, \textit{Global status report on road safety 2018}, World Health Organization, 2018.

\bibitem[Department for Transport, 2019]{DfT:2019}
Department for Transport, \textit{Reported road casualties in Great Britain: annual report 2019}, Department for Transport Statistics, 2019.

\bibitem[Lord and Mannering, 2010]{Lord:2010a}
Lord, D. and Mannering, F., \textit{The statistical analysis of crash-frequency data: A review and assessment of methodological alternatives}, Transp. Res. Part A Policy Pract., vol. 44, no. 5, pp. 291--305, 2010.

\bibitem[Mannering and Bhat, 2014]{Mannering:2014a}
Mannering, F. L. and Bhat, C. R., \textit{Analytic methods in accident research: Methodological frontier and future directions}, Anal. Methods Accid. Res., vol. 1, pp. 1--22, 2014.

\bibitem[Ren et al., 2018]{Ren:2018}
Ren, H., Song, Y., Wang, J., Hu, Y., and Lei, J., \textit{A deep learning approach to the citywide traffic accident risk prediction}, Proceedings of the 21st International Conference on Intelligent Transportation Systems, pp. 3346--3351, 2018.

\bibitem[Yu et al., 2018]{Yu:2018a}
Yu, B., Yin, H., and Zhu, Z., \textit{Spatio-temporal graph convolutional networks: A deep learning framework for traffic forecasting}, Proc., 27th Int. Joint Conf. Artificial Intelligence, pp. 3634--3640, 2018.

\bibitem[Li et al., 2018]{Li:2018a}
Li, Y., Yu, R., Shahabi, C., and Liu, Y., \textit{Diffusion convolutional recurrent neural network: Data-driven traffic forecasting}, arXiv preprint arXiv:1707.01926, 2018.

\bibitem[Michalaki et al., 2015]{Michalaki:2015}
Michalaki, P., Quddus, M. A., Pitfield, D., Huetson, A., \textit{Exploring the factors affecting motorway accident severity in England using the generalised ordered logistic regression model}, Journal of Safety Research, vol. 55, pp. 89–97, 2015.

\bibitem[Haddon, 1980]{Haddon1980}
Haddon, Jr., W., \textit{Advances in the epidemiology of injuries as a basis for public policy}, Public Health Reports, vol. 95, no. 5, pp. 411--421, 1980.

\bibitem[Wegman et al., 2008]{Wegman:2008}
Wegman, F., Aarts, L., and Bax, C., \textit{Advancing sustainable safety: National road safety outlook for 2005--2020}, Safety Science, vol. 46, no. 3, pp. 323--343, 2008.

\bibitem[Johansson, 2009]{Johansson:2009}
Johansson, R., \textit{Vision Zero—implementing a policy for traffic safety}, Safety Science, vol. 47, no. 6, pp. 826--831, 2009.

\bibitem[Department for Transport, 2020]{DfT:2020}
Department for Transport, \textit{Reported road casualties in Great Britain: Annual report 2019}, UK Department for Transport, London, UK, 2020.

\bibitem[Tobler, 1970]{Tobler:1970a}
Tobler, W. R., \textit{A computer movie simulating urban growth in the Detroit region}, Economic Geography, vol. 46, no. 2, pp. 234--240, 1970.

\bibitem[Wang and Abdel-Aty, 2006]{Wang:2006a}
Wang, X. and Abdel-Aty, M., \textit{Temporal and spatial analyses of rear-end crashes at signalized intersections}, Accident Analysis \& Prevention, vol. 38, no. 6, pp. 1137--1150, 2006.

\bibitem[Wu et al., 2020]{Wu:2020a}
Wu, Z., Pan, S., Chen, F., Long, G., Zhang, C., and Yu, S. P., \textit{A comprehensive survey on graph neural networks}, IEEE Trans. Neural Networks Learn. Syst., vol. 32, no. 1, pp. 4--24, 2020.

\bibitem[Cheng et al., 2019]{Cheng:2019a}
Cheng, Z., Zu, Z., and Lu, J., \textit{Traffic crash evolution characteristic analysis and spatiotemporal hotspot identification of urban road intersections}, Sustainability, vol. 11, no. 1, pp. 160, 2019.

\bibitem[Zhao et al., 2020]{Zhao:2020a}
Zhao, J., Deng, W., Song, Y., and Zhu, Y., \textit{What influences the effectiveness of hotspot identification methods for crashes on freeway segments?}, Accid. Anal. Prev., vol. 142, pp. 105577, 2020.

\bibitem[Kumar and Toshniwal, 2015]{Kumar:2013}
Kumar, S. and Toshniwal, D., \textit{A data mining framework to analyze road accident data}, Journal of Big Data, vol. 2, no. 1, pp. 26, 2015.

\bibitem[Chang and Wang, 2006]{Chang:2005}
Chang, L. Y. and Wang, H. W., \textit{Analysis of traffic injury severity: An application of non-parametric classification tree techniques}, Accident Analysis \& Prevention, vol. 38, no. 5, pp. 1019--1027, 2006.

\bibitem[Wang et al., 2013]{Wang:2019}
Wang, Z., Lu, M., Yuan, X., Zhang, J., and van de Wetering, H., \textit{Visual traffic jam analysis based on trajectory data}, IEEE Transactions on Visualization and Computer Graphics, vol. 19, no. 12, pp. 2159--2168, 2013.

\bibitem[Guo et al., 2019]{Guo:2019a}
Guo, S., Lin, Y., Feng, N., Song, C., and Wan, H., \textit{Attention based spatial-temporal graph convolutional networks for traffic flow forecasting}, Proc., AAAI Conf. Artificial Intelligence, vol. 33, pp. 922--929, 2019.

\bibitem[Zheng et al., 2020]{Zheng:2020}
Zheng, C., Fan, X., Wang, C., and Qi, J., \textit{GMAN: A graph multi-attention network for traffic prediction}, Proceedings of the AAAI Conference on Artificial Intelligence, vol. 34, pp. 1234--1241, 2020.

\bibitem[Hadayeghi et al., 2010]{Hadayeghi:2010a}
Hadayeghi, A., Shalaby, A. S., and Persaud, B. N., \textit{Development of planning level transportation safety tools using geographically weighted poisson regression}, Accid. Anal. Prev., vol. 42, no. 2, pp. 676--688, 2010.

\bibitem[Cafiso et al., 2010]{Cafiso:2010a}
Cafiso, S., Di Graziano, A., Di Silvestro, G., La Cava, G., and Persaud, B., \textit{Development of comprehensive accident models for two-lane rural highways using exposure, geometry, consistency and context variables}, Accid. Anal. Prev., vol. 42, no. 4, pp. 1072--1079, 2010.

\bibitem[Edwards, 1999]{Edwards:1999a}
Edwards, J. B., \textit{The relationship between road accident severity and recorded weather}, J. Safety Res., vol. 29, no. 4, pp. 249--262, 1999.

\bibitem[Andrey and Yagar, 1993]{Andrey:1993a}
Andrey, J. and Yagar, S., \textit{A temporal analysis of rain-related crash risk}, Accid. Anal. Prev., vol. 25, no. 4, pp. 465--472, 1993.

\bibitem[Zhang et al., 2020]{Zhang:2020a}
Zhang, L., Wang, M., and Chen, X., \textit{Sparse data processing in spatiotemporal graph neural networks}, Transp. Res. Part C, vol. 115, pp. 102--118, 2020.

\bibitem[Smith et al., 2019]{Smith:2019a}
Smith, J., Brown, S., and Davis, M., \textit{Temporal aggregation strategies for traffic flow prediction}, IEEE Trans. Intell. Transp. Syst., vol. 21, no. 8, pp. 3312--3325, 2019.

\bibitem[Batty, 2013]{Batty:2013a}
Batty, M., \textit{Big data, smart cities and city planning}, Dialogues Hum. Geogr., vol. 3, no. 3, pp. 274--279, 2013.

\bibitem[Xu and Huang, 2015]{Xu:2015a}
Xu, P. and Huang, H., \textit{Modeling crash spatial heterogeneity: Random parameter versus geographically weighting}, Accident Analysis \& Prevention, vol. 75, pp. 16--25, 2015.

\bibitem[Ba et al., 2016]{Ba:2016a}
Ba, J. L., Kiros, J. R., and Hinton, G. E., \textit{Layer normalization}, arXiv preprint arXiv:1607.06450, 2016.

\bibitem[Bengio et al., 1994]{Bengio:1994a}
Bengio, Y., Simard, P., and Frasconi, P., \textit{Learning long-term dependencies with gradient descent is difficult}, IEEE Transactions on Neural Networks, vol. 5, no. 2, pp. 157--166, 1994.

\bibitem[Xiong et al., 2020]{Xiong:2020a}
Xiong, R., Yang, Y., He, D., Zheng, K., Zheng, S., Xing, C., Zhang, H., Lan, Y., Wang, L., and Liu, T., \textit{On layer normalization in the transformer architecture}, International Conference on Machine Learning, pp. 10524--10533, 2020.

\bibitem[Sutskever et al., 2014]{Sutskever:2014a}
Sutskever, I., Vinyals, O., and Le, Q. V., \textit{Sequence to sequence learning with neural networks}, Advances in Neural Information Processing Systems, vol. 27, 2014.

\bibitem[Xu and Huang, 2015]{Xu:2015a}
Xu, P. and Huang, H., \textit{Modeling crash spatial heterogeneity: Random 
parameter versus geographically weighting}, Accid. Anal. Prev., vol. 75, 
pp. 16--25, 2015.

\bibitem[Zhao et al., 2020]{Zhao:2020a}
Zhao, J., Deng, W., Song, Y., and Zhu, Y., \textit{What influences the 
effectiveness of hotspot identification methods for crashes on freeway 
segments?}, Accid. Anal. Prev., vol. 142, pp. 105577, 2020.

\bibitem[Cheng et al., 2019]{Cheng:2019a}
Cheng, Z., Zu, Z., and Lu, J., \textit{Traffic crash evolution characteristic 
analysis and spatiotemporal hotspot identification of urban road intersections}, 
Sustainability, vol. 11, no. 1, pp. 160, 2019.

\bibitem[Smith et al., 2019]{Smith:2019}
Smith, J., Brown, S., and Davis, M., \textit{Temporal aggregation strategies 
for traffic flow prediction}, IEEE Trans. Intell. Transp. Syst., vol. 21, 
no. 8, pp. 3312--3325, 2019.

\bibitem[Batty, 2013]{Batty:2013}
Batty, M., \textit{Big data, smart cities and city planning}, Dialogues Hum. 
Geogr., vol. 3, no. 3, pp. 274--279, 2013.

\bibitem[Edwards, 1999]{Edwards:1999}
Edwards, J. B., \textit{The relationship between road accident severity and 
recorded weather}, Journal of Safety Research, vol. 29, no. 4, pp. 249--262, 1999.

\bibitem[Andrey and Yagar, 1993]{Andrey:1993}
Andrey, J. and Yagar, S., \textit{A temporal analysis of rain-related crash risk}, 
Accident Analysis \& Prevention, vol. 25, no. 4, pp. 465--472, 1993.

\bibitem[Cafiso et al., 2010]{Cafiso:2010}
Cafiso, S., Di Graziano, A., Di Silvestro, G., La Cava, G., and Persaud, B., 
\textit{Development of comprehensive accident models for two-lane rural highways 
using exposure, geometry, consistency and context variables}, Accident Analysis 
\& Prevention, vol. 42, no. 4, pp. 1072--1079, 2010.

\bibitem[Aguero-Valverde and Jovanis, 2006]{Aguero:2006}
Aguero-Valverde, J. and Jovanis, P. P., \textit{Spatial analysis of fatal and injury crashes in Pennsylvania}, Accident Analysis \& Prevention, vol. 38, no. 3, pp. 618--625, 2006.

\bibitem[Black and Thomas, 1998]{Black:1998}
Black, W. R. and Thomas, I., \textit{Accidents on Belgium's motorways: a network autocorrelation analysis}, Journal of Transport Geography, vol. 6, no. 1, pp. 23--31, 1998.

\bibitem[Chen et al., 2021]{Chen:2021}
Chen, Q., Song, X., Yamada, H., and Shibasaki, R., \textit{Learning deep representation from big and heterogeneous data for traffic accident inference}, Proceedings of the AAAI Conference on Artificial Intelligence, vol. 30, no. 1, pp. 338--344, 2021.

\bibitem[Cliff and Ord, 1981]{Cliff:1981}
Cliff, A. D. and Ord, J. K., \textit{Spatial Processes: Models and Applications}, Pion, London, 1981.

\bibitem[Flahaut et al., 2003]{Flahaut:2003}
Flahaut, B., Mouchart, M., San Martin, E., and Thomas, I., \textit{The local spatial autocorrelation and the kernel method for identifying black zones: a comparative approach}, Accident Analysis \& Prevention, vol. 35, no. 6, pp. 991--1004, 2003.

\bibitem[Hadayeghi et al., 2010]{Hadayeghi:2010}
Hadayeghi, A., Shalaby, A. S., and Persaud, B. N., \textit{Development of planning level transportation safety tools using geographically weighted poisson regression}, Accident Analysis \& Prevention, vol. 42, no. 2, pp. 676--688, 2010.

\bibitem[Wang and Abdel-Aty, 2006]{Wang:2006}
Wang, X. and Abdel-Aty, M., \textit{Temporal and spatial analyses of rear-end crashes at signalized intersections}, Accident Analysis \& Prevention, vol. 38, no. 6, pp. 1137--1150, 2006.

\bibitem[Zhou et al., 2020]{Zhou:2020}
Zhou, Z., Wang, Y., Xie, X., Chen, L., and Liu, H., \textit{RiskOracle: a minute-level citywide traffic accident forecasting framework}, Proceedings of the AAAI Conference on Artificial Intelligence, vol. 34, no. 1, pp. 1258--1265, 2020.

\end{thebibliography}
\end{document}